\documentclass[10pt,twocolumn,letterpaper]{article}

\usepackage[final]{iccv}      

\usepackage{times}
\usepackage{epsfig}
\usepackage{graphicx}
\usepackage{amsmath}
\usepackage{amssymb}
\usepackage{wasysym}
\usepackage{bm}
\usepackage{booktabs}
\usepackage{comment}
\usepackage{array}
\usepackage{multirow} 
\usepackage{tabularx}
\usepackage[a4paper, margin=1in]{geometry}

\definecolor{iccvblue}{rgb}{0.21,0.49,0.74}
\usepackage[pagebackref,breaklinks,colorlinks,allcolors=iccvblue]{hyperref}

\newcolumntype{L}[1]{>{\raggedright\arraybackslash}m{#1}}
\newcolumntype{C}[1]{>{\centering\arraybackslash}m{#1}}
\newcolumntype{R}[1]{>{\raggedleft\arraybackslash}m{#1}}
\newcolumntype{+}{>{\global\let\currentrowstyle\relax}}
\newcolumntype{^}{>{\currentrowstyle}}

\newcommand{\bfx}{\ensuremath{{\mathbf{x}}}}

\newcommand{\bfc}{\ensuremath{{\mathbf{c}}}}
\newcommand{\bfu}{\ensuremath{{\mathbf{u}}}}
\newcommand{\bfv}{\ensuremath{{\mathbf{v}}}}
\newcommand{\bfs}{\ensuremath{{\mathbf{s}}}}

\newcommand{\bfU}{\ensuremath{{\mathbf{U}}}}

\newcommand{\bfV}{\ensuremath{{\mathbf{V}}}}

\def\paperID{11005} 
\def\confName{ICCV}
\def\confYear{2025}

\title{\makebox[\textwidth][c]{Oneta: Multi-Style Image Enhancement Using Eigentransformation Functions}}

\author{
\makebox[0.9\textwidth][c]{Jiwon Kim$^{1}$\thanks{Equal contribution.} , Soohyun Hwang$^{1}$\footnotemark[1] , Dong-O Kim$^2$, Changsu Han$^2$, Min Kyu Park$^2$, Chang-Su Kim$^1$}\\
\makebox[0.9\textwidth][c]{Korea University$^1$, Department of Camera Innovation Group, Samsung Electronics$^2$}\\
\makebox[0.9\textwidth][c]{\tt\footnotesize \{jwkim, shhwang\}@mcl.korea.ac.kr} \\
\makebox[0.9\textwidth][c]{\tt\footnotesize \{dongo.kim, liebein.han, minkyu7.park\}@samsung.com, changsukim@korea.ac.kr}
}

\begin{document}
\maketitle
\begin{abstract}
The first algorithm, called Oneta, for a novel task of multi-style image enhancement is proposed in this work. Oneta uses two point operators sequentially: intensity enhancement with a transformation function (TF) and color correction with a color correction matrix (CCM). This two-step enhancement model, though simple, achieves a high performance upper bound. Also, we introduce eigentransformation function (eigenTF) to represent TF compactly. The Oneta network comprises Y-Net and C-Net to predict eigenTF and CCM parameters, respectively. To support $K$ styles, Oneta employs $K$ learnable tokens. During training, each style token is learned using image pairs from the corresponding dataset. In testing, Oneta selects one of the $K$ style tokens to enhance an image accordingly. Extensive experiments show that the single Oneta network can effectively undertake six enhancement tasks --- retouching, image signal processing, low-light image enhancement, dehazing, underwater image enhancement, and white balancing --- across 30 datasets. 
\end{abstract}

\section{Introduction}

Image enhancement is a low-level vision task to improve the perceptual quality, visibility, and appearance of an image. It is essential for subsequent downstream tasks, including object detection, medical imaging, and autonomous driving. Image enhancement can be divided into several sub-tasks, depending on the objectives, such as retouching, image signal processing (ISP), low-light image enhancement (LLIE), dehazing, underwater image enhancement (UIE), and white balancing (WB). 

There are many traditional enhancement algorithms, including point operators \cite{DIP_BOOK_Jain, gonzalez2007dip,lee2013contrast}. Recently, deep learning algorithms have been developed, based on convolutional neural network (CNN) or vision transformer (ViT), for retouching \cite{zheng2022lut, yang2022adaint, yang2022seplut, kim2020pienet, kim2024image, zhang2021star}, ISP \cite{xing2021invertible, kim2023paramisp}, LLIE \cite{wang2019underexposed, cai2023retinexformer, xu2022snr, wang2023low}, dehazing \cite{ye2022perceiving, zheng2023curricular, guo2022image, qiu2023mb}, UIE \cite{li2021underwater, tang2023underwater, peng2023u, zhao2024wavelet}, and WB \cite{afifi2018semantic, afifi2020deep}, respectively. All these algorithms were designed to undertake each specific task only. Meanwhile, attempts have been made to perform multiple image processing tasks \cite{tu2022maxim,jiang2024fast, chen2021pre, zamir2022restormer, liang2022recurrent, li2023efficient, liang2021swinir, zhang2022accurate, wang2022uformer}. These multi-task algorithms, however, are mainly for image restoration rather than enhancement. Also, they should be trained separately for each task or dataset, or they require extra network components for multi-tasking. 

\begin{figure}[t]
\centering 
\includegraphics[width=1\linewidth]{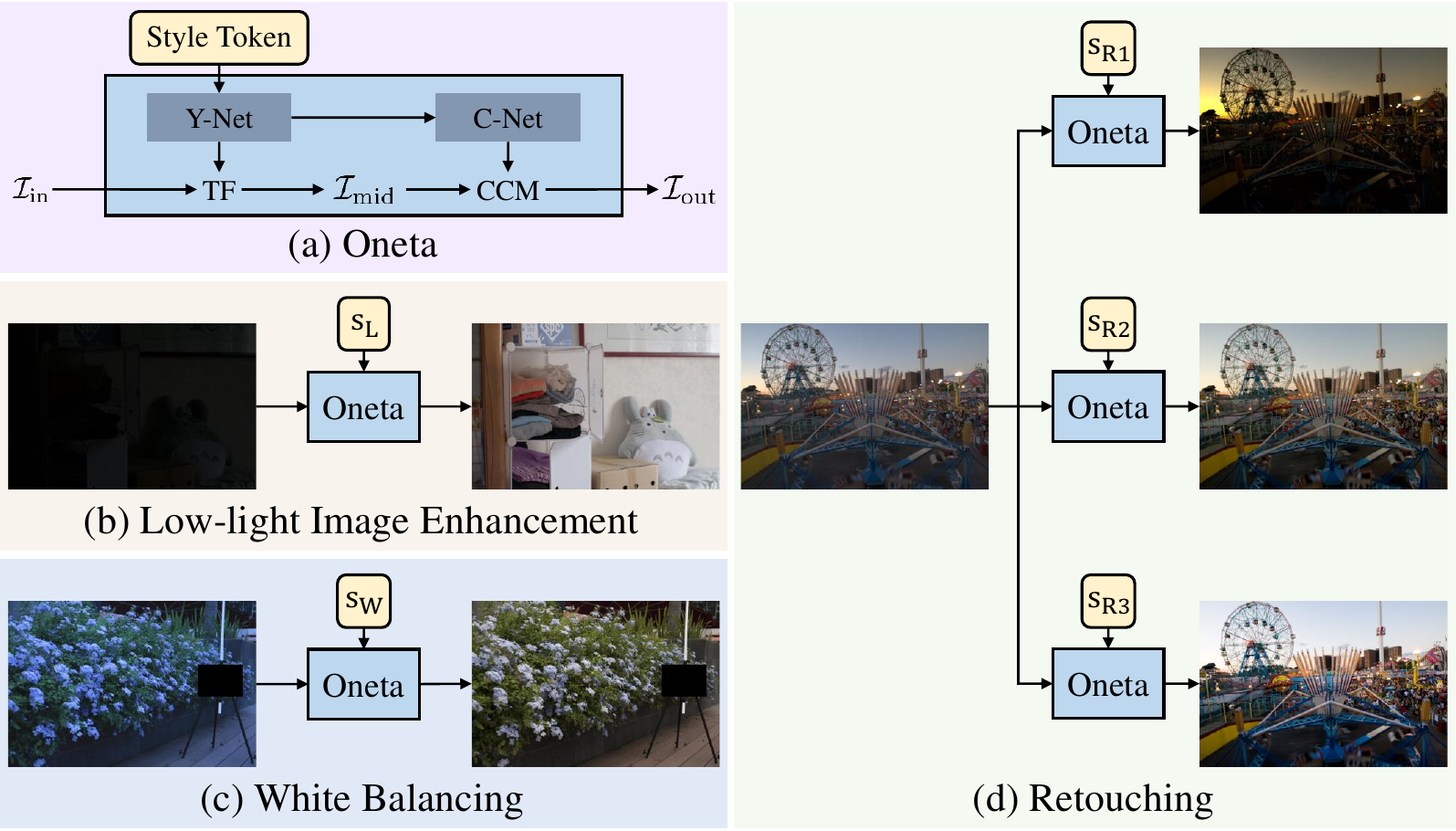}
\caption{Given an input image $\mathcal{I}_\mathrm{in}$ and a style token, the proposed Oneta algorithm generates an output image $\mathcal{I}_\mathrm{out}$ in the desired style. Oneta can retouch an identical image in different styles and perform a variety of tasks, including LLIE and WB, simply by changing the style token.} 
\label{fig:oneta}
\end{figure}

In this paper, we propose a novel algorithm called Oneta\footnote{It is named after `one network to enhance them all.' However, `all' should be interpreted as all the datasets used for Oneta training.} for multi-style image enhancement. Figure~\ref{fig:oneta} illustrates the Oneta algorithm, which takes an input image and a style token to generate an output image in the desired style. In this example, Oneta retouches an identical image in three different styles. Also, it copes with the different tasks of LLIE and WB as well. Remarkably, Oneta can achieve this multi-style enhancement using a single network without retraining or extra components, such as multi-tails or multi-heads \cite{chen2021pre}. It just requires the switching of a style token depending on the desired style.

For reliable multi-style enhancement, Oneta uses only two point operators sequentially, as shown in Figure~\ref{fig:oneta}(a): intensity enhancement through a transformation function (TF) and color correction through a color correction matrix (CCM). It is shown via quadratic programming (QP) \cite{boyd2004convex} that this two-step enhancement model, though simple, has a relatively high performance upper bounds. Moreover, to further improve the reliability of enhancement, we introduce the notion of eigentransformation function (eigenTF) and represent a TF compactly with $M$ eigenTF coefficients. This is achieved by the best rank-$M$ approximation \cite{y2015_SVD_Hopcroft} of the optimum TFs, derived via QP. Consequently, Oneta should predict only a few parameters to enhance an image. 

The Oneta network in Figure~\ref{fig:oneta}(a) consists of Y-Net and C-Net, both based on ViT  \cite{dosovitskiy2020image}. Y-Net predicts eigenTF coefficients, and C-Net does CCM parameters. In addition to image tokens, Oneta takes an extra (\ie style) token to enhance an image in the corresponding style. To support $K$ styles, Oneta employs $K$ learnable tokens. In the training phase, each style token is learned using image pairs from the corresponding dataset. In the inference phase, Oneta selects one of the $K$ style tokens to enhance an image accordingly. 

This work has the following major contributions:
\begin{itemize}
\itemsep1mm
\item We propose a novel task of multi-style enhancement, which aims to undertake diverse enhancement tasks and improve images in various styles using a single network without retraining. 
\item To enable multi-style enhancement, we implement the Oneta network composed of Y-Net and C-Net, based on ViT, and introduce switchable style tokens. 
\item For reliable enhancement, we develop a simple two-step enhancement model and show that its performance upper bound is relatively high. Also, we introduce the notion of eigenTF to perform the first step even more reliably. 
\item Extensive experiments demonstrate that Oneta can effectively cope with six image enhancement tasks of retouching, ISP, LLIE, dehazing, UIE, and WB across 30 datasets, as illustrated in Figure~\ref{fig:image_enhancements_examples}. 
\end{itemize}

\section{Related Work}

\subsection{Global Image Enhancement}

Images can be enhanced by global or local techniques. Global enhancement uses point operators \cite{DIP_BOOK_Jain}, which apply the same transformations to all pixels in an image to improve brightness, contrast, or color balance. Even though it has more design constraints than spatially adaptive local enhancement, global enhancement is computationally simpler and yields more reliable results in general. 

Gamma correction and histogram equalization are well-known point operators \cite{gonzalez2007dip}. More sophisticated methods for determining TFs or lookup tables (LUTs) also have emerged. For example, Lee \etal \cite{lee2013contrast} derived a TF to emphasize the intensity differences between adjacent pixels. Recently, deep networks have been developed to predict LUTs effectively. Zheng \etal \cite{zheng2022lut} first proposed to learn 3D LUTs. Yang \etal \cite{yang2022adaint} learned predicting image-adap\-tive sampling points to increase the nonlinearity of LUTs. Also, Yang \etal \cite{yang2022seplut} employed a cascade of 1D and 3D LUTs to improve the computational efficiency. 

To enhance images reliably in various styles upon user requests, the proposed Oneta adopts two simple global operators: one for intensity enhancement and the other for color correction. Moreover, we propose the notion of eigenTFs to reduce the number of parameters that should be predicted, thereby improving the reliability of enhancement further. 

\begin{figure}[t]
\centering 
\includegraphics[width=1\linewidth]{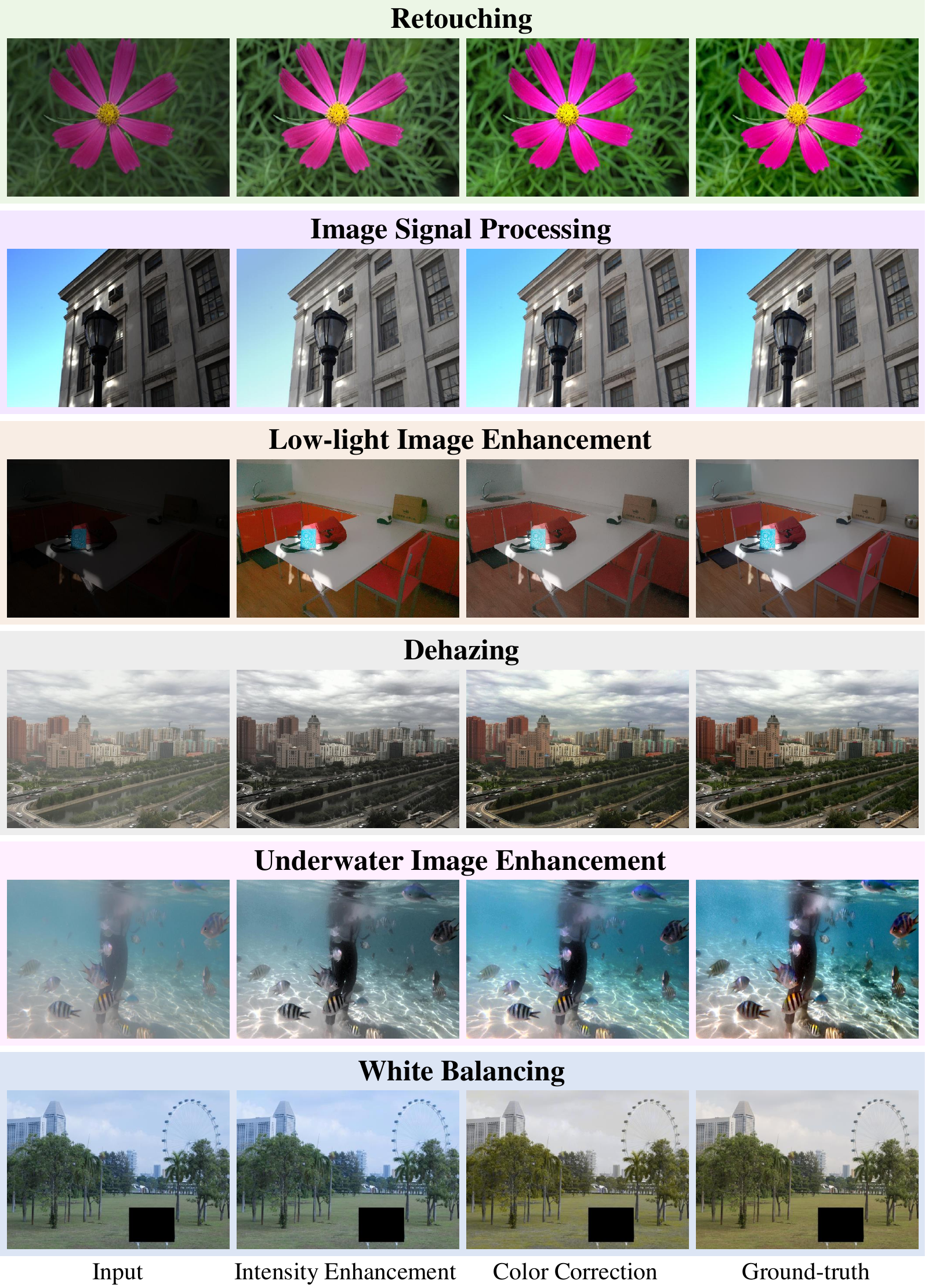}
\caption{Oneta can cope with retouching, ISP, LLIE, dehazing, UIE, and WB by performing intensity enhancement and color correction sequentially.} 
\label{fig:image_enhancements_examples}
\end{figure}

\subsection{Multi-Task Enhancement} 

Deep learning has been adopted for increasingly many enhancement tasks, including retouching \cite{kim2020pienet, yang2022adaint, yang2022seplut, kim2024image}, ISP \cite{xing2021invertible, kim2023paramisp}, LLIE \cite{wang2019underexposed, xu2022snr, cai2023retinexformer}, dehazing \cite{ye2022perceiving, zheng2023curricular}, UIE \cite{li2021underwater, tang2023underwater, zhao2024wavelet}, and WB \cite{afifi2018semantic, afifi2020deep}. Different strategies and networks have been developed specifically for these various tasks. However, these tasks share the common goal of improving image quality by enhancing pixel intensities and colors, as shown in Figure~\ref{fig:image_enhancements_examples}. 

\begin{figure*}[t]
\centering
\includegraphics[width=1\linewidth]{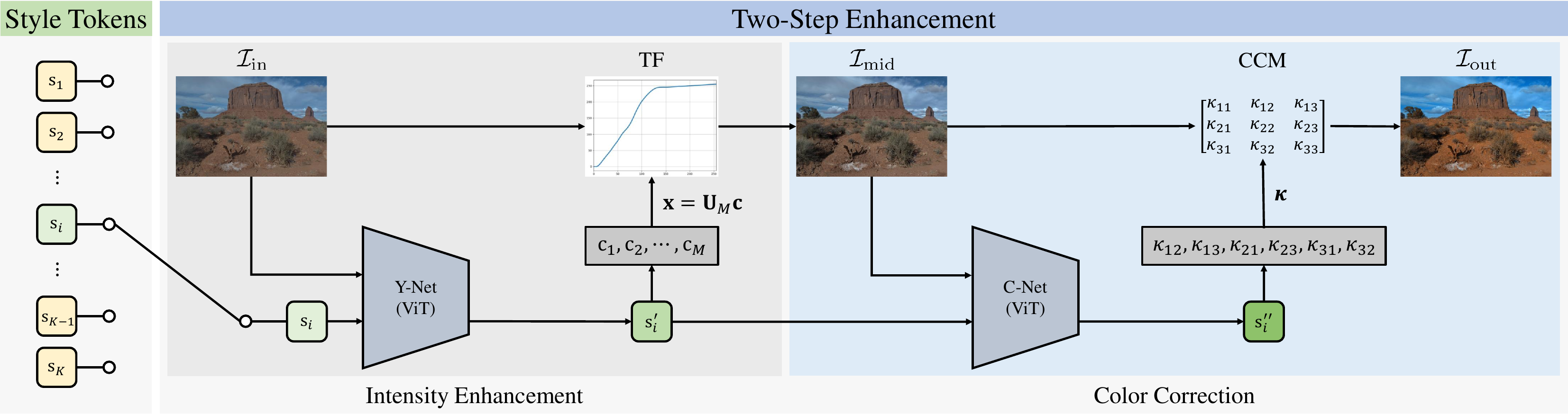}
\caption{Overview of the proposed Oneta algorithm: In Oneta, a user selects one of $K$ style tokens. First, to perform intensity enhancement, Y-Net attends the selected style token $\bfs_i$ to embedding vectors from an input image $\mathcal{I}_\mathrm{in}$. The attended style token $\bfs_i'$ is used to reconstruct the TF $\bfx$, which transforms $\mathcal{I}_\mathrm{in}$ into the intermediate image $\mathcal{I}_\mathrm{mid}$. Second, for color correction, C-Net further processes $\bfs_i'$ with $\mathcal{I}_\mathrm{mid}$ to generate  $\bfs_i''$. The CCM, predicted using $\bfs_i''$, is then employed to transform $\mathcal{I}_\mathrm{mid}$ into the final output image $\mathcal{I}_\mathrm{out}$.}
\label{fig:overview}
\vspace*{-0.2cm}
\end{figure*}

Attempts have been made to develop encoder-decoder networks that can perform multiple image processing tasks \cite{tu2022maxim, jiang2024fast}. For instance, Jiang \etal \cite{jiang2024fast} covered nine tasks of deraining, motion/defocus deblurring, denoising, dehazing, desnowing, super-resolution, UIE, and LLIE across 31 datasets. However, these algorithms focus more on image restoration than enhancement. Moreover, they should be trained separately for each task or dataset. In contrast, Chen \etal \cite{chen2021pre} proposed learning multiple tasks simultaneously. They used multi-heads and multi-tails to perform three restoration tasks of denoising, deraining, and super-resolution with a shared transformer body.

In this paper, we propose a universal network, Oneta, to perform multiple enhancement tasks without any retraining or additional components, such as multi-heads/tails. For each task or dataset, Oneta just requires the switching of a style token, as shown in Figure~\ref{fig:oneta}. The current implementation of Oneta can handle six enhancement tasks of retouching, ISP, LLIE, dehazing, UIE, and WB across 30 datasets.

\subsection{Vision Transformer}

Initially applied to image classification, ViT \cite{dosovitskiy2020image} has been successfully employed for image processing tasks, including LLIE \cite{xu2022snr,  cai2023retinexformer, wang2023low}, dehazing \cite{guo2022image, qiu2023mb},  UIE \cite{peng2023u, zhao2024wavelet}, deblurring \cite{tsai2022stripformer}, denoising \cite{song2022tempformer}, and super-resolution \cite{li2023feature, zhang2024realviformer}. Moreover, several attempts have been made to perform multiple image processing tasks with transformers, but still limited to image restoration, such as deblurring, denoising, and super-resolution \cite{liang2022recurrent, li2023efficient, liang2021swinir, zhang2022accurate, wang2022uformer, zamir2022restormer}. Meanwhile, Zhang \etal \cite{zhang2021star} developed a two-branch transformer module to replace the CNN backbones of conventional enhancement methods for retouching, LLIE, and WB.  

In ViT \cite{dosovitskiy2020image}, an extra token, in addition to image tokens, is used for the classification task. Some methods use more extra tokens to accomplish their objectives. Touvron \etal \cite{touvron2021training} used a distillation token, as well as a class token, to facilitate teacher-student knowledge distillation. For semantic segmentation, Xu \etal \cite{xu2022multi} employed multiple class tokens simultaneously to learn class-specific object localization maps. 

In this paper, we find new usage for this extra token. During training, we learn $K$ extra tokens, called style tokens, where $K$ is the number of enhancement styles that Oneta supports. In testing, a single style token is selected from them and used to enhance an image accordingly. Thus, we can support multi-style enhancement with a single network.

\section{Proposed Algorithm}
Figure~\ref{fig:overview} is an overview of the proposed Oneta algorithm for multi-style image enhancement. 

\subsection{Two-Step Enhancement}
For reliable multi-style image enhancement, we use a simple enhancement model that has only two steps: intensity enhancement and color correction. Both steps employ global point operators, so an output pixel $(r_\mathrm{out}, g_\mathrm{out}, b_\mathrm{out})$ is determined by the corresponding input pixel $(r_\mathrm{in}, g_\mathrm{in}, b_\mathrm{in})$ only. 
Let $(r_\mathrm{mid}, g_\mathrm{mid}, b_\mathrm{mid})$ denote the intermediate pixel after the first step of intensity enhancement. Also, let $y_\mathrm{in}$, $y_\mathrm{mid}$, and $y_\mathrm{out}$ be the intensities of the input, intermediate, and output pixels, respectively. 

\vspace*{0.1cm}
\noindent\textbf{Intensity enhancement:} We use a TF, denoted by a column vector
\begin{equation}\label{eq:TF}
\textstyle
\mathbf{x}=[x_0, x_1, \ldots, x_{255}]^t    
\end{equation}
where $x_k$ is the enhanced intensity when the original intensity is $k$. Thus, we enhance $y_\mathrm{in}$ to $y_\mathrm{mid}=x_{y_\mathrm{in}}$. Then, the intermediate pixel is given by 
\begin{equation}\label{eq:rgb_scaling}
\textstyle
 (r_\mathrm{mid}, g_\mathrm{mid}, b_\mathrm{mid}) =  \frac{y_\mathrm{mid}}{y_\mathrm{in}} (r_\mathrm{in}, g_\mathrm{in}, b_\mathrm{in}).
\end{equation}

\vspace*{0.1cm}
\noindent\textbf{Color correction:} We multiply a CCM and the intermediate pixel to obtain the output pixel,
\begin{equation}\label{eq:CCM}
\textstyle
    \begin{bmatrix}
     r_\mathrm{out} \\
     g_\mathrm{out} \\
     b_\mathrm{out} 
    \end{bmatrix} 
    =
   \begin{bmatrix}
    \kappa_{11} & \kappa_{12} & \kappa_{13}\\
    \kappa_{21} & \kappa_{22} & \kappa_{23}\\
    \kappa_{31} & \kappa_{32} & \kappa_{33}
    \end{bmatrix} 
    \begin{bmatrix}
     r_\mathrm{mid} \\
     g_\mathrm{mid} \\
     b_\mathrm{mid} 
    \end{bmatrix}
\end{equation}
where we adopt the row sum constraints, $\sum_{j} \kappa_{ij} = 1$ for each $i$ \cite{bianco2013color}.

\subsection{Optimum TF and CCM}
Despite its simplicity, the two-step enhancement model achieves high performance upper bounds, as shown in Section~\ref{sec:exp}. Given a pair of input and ground-truth (GT) images, an upper bound can be obtained by finding the optimum TF and CCM. Let us describe how to derive these optimum TF and CCM via QP \cite{boyd2004convex}. 

Let $\mathbf{y}_\mathrm{in}$ be the column vector containing intensities in an input image $\mathcal{I}_\mathrm{in}$. Specifically, its $i$th element $y_\mathrm{in}^i$ is the intensity of the $i$th pixel in the raster scan order. To find the optimum TF $\mathbf{x}^*$, we form an $N\times 256$ matrix $\mathbf{A}=[\mathbf{a}_1,\mathbf{a}_2, \cdots, \mathbf{a}_N]^t$,
where $N$ is the number of pixels in an image. Also, $\mathbf{a}_i$ is a one-hot vector whose $y_\mathrm{in}^i$-th element is the only 1. Then, the intermediate intensity vector is obtained by multiplying $\mathbf{A}$ with the TF $\mathbf{x}$ in \eqref{eq:TF},   
\begin{equation}\label{eq:intensity_enhancement} 
\textstyle
 \mathbf{y}_\mathrm{mid} = \mathbf{A} \mathbf{x}.
\end{equation}
It is desired that this $\mathbf{y}_\mathrm{mid}$ is as close to the GT $\mathbf{y}_\mathrm{GT}$ as possible. Hence, the optimum TF $\mathbf{x}^*$ can be obtained by solving a QP problem; 
\begin{equation}\label{eq:QP_TF}
\textstyle
    \mathbf{x}^* =  {\underset{\mathbf{x}}{\arg\min}} \|\mathbf{A} \mathbf{x}-\mathbf{y}_\mathrm{GT}\|^2
\end{equation}
subject to the monotonicity constraints of the TF 
$0\leq x_0 \leq x_1 \leq \cdots \leq x_{255} \leq 255$.

After obtaining $\mathbf{x}^*$, we reconstruct the optimum intermediate image $\mathcal{I}_\mathrm{mid}^*$ via \eqref{eq:intensity_enhancement} and \eqref{eq:rgb_scaling}. Then, as detailed in the supplement, we construct a $3N\times 9$ matrix $\mathbf{B}$ from $\mathcal{I}_\mathrm{mid}^*$ so that the product $\mathbf{B}\bm{\kappa}$ represent the color-corrected RGB values. Here, $\bm{\kappa}$ is a 9-dimensional vector reshaped from the CCM in \eqref{eq:CCM}. Also, we form a vector $\mathbf{z}_\mathrm{GT}$ composed of the $3N$ RGB values in the GT image $\mathcal{I}_\mathrm{GT}$. Then, we determine the optimum CCM $\bm{\kappa}^*$ by solving another QP problem;
\begin{equation} \label{eq:QP_CCM}
\textstyle
\bm{\kappa}^* = {\underset{\bm{\kappa}}{\arg\min}} \| \mathbf{B}\bm{\kappa} - \mathbf{z}_\mathrm{GT} \|^2
\end{equation}
subject to the row sum constraints
\begin{equation}\label{eq:row_sum}
\textstyle
    \sum_{j=1}^3 \kappa_{ij} = 1 \mbox{ for each } i=1,2,3.
\end{equation}

\subsection{EigenTF}
In the inference phase, without the GT, we should predict the TF and CCM to enhance an input image. Note that the TF in \eqref{eq:TF} has 256 parameters, whereas the CCM in \eqref{eq:CCM} needs only 6 parameters because the diagonal elements are determined by the 6 non-diagonal ones via \eqref{eq:row_sum}. For reliable image enhancement, it is beneficial to reduce the number of parameters to be predicted. Hence, we propose  eigenTFs to represent TFs compactly.    

First, we construct a TF matrix ${\mathbf X}=[{\mathbf x}_1^*, {\mathbf x}_2^*, \cdots, {\mathbf x}_L^*]$ consisting of the optimum TFs, obtained by \eqref{eq:QP_TF}, for all $L$ training images. Then, we apply singular value decomposition (SVD) to $\mathbf X$,
\begin{equation}\label{eq:svd}
    \textstyle
    \mathbf{X} = \mathbf{U} \mathbf{\Sigma} \mathbf{V}^t
\end{equation}
where $\mathbf{U} = [\bfu_1, \cdots, \bfu_{255}]$ and $\bfV = [\bfv_1, \cdots, \bfv_L]$ are orthogonal matrices. Also, $\mathbf{\Sigma}$ is a diagonal matrix of singular values $\sigma_1 \geq \sigma_2 \geq \cdots \geq \sigma_r > 0$, where $r$ is the rank of $\mathbf{X}$. Then, we can approximate each $\bfx_i^*$ by linearly combining the first $M$ vectors $\bfu_1, \cdots, \bfu_M$ in $\mathbf{U}$, 
\begin{equation} \label{eq:x_approx}
\textstyle
\tilde{\bfx}_i^* = \bfU_M \bfc_i = [\bfu_1, \cdots, \bfu_M] \bfc_i,
\end{equation}
which is known to be the best rank-$M$ approximation~\cite{y2015_SVD_Hopcroft}.

We call $\bfu_1, \cdots, \bfu_M$ in \eqref{eq:x_approx} as eigenTFs, and the space spanned by them as the eigenTF space. Given a TF $\bfx$, we project it onto the eigenTF space to obtain the low-rank approximation
\begin{equation}\label{eq:backward}
\textstyle
    \tilde{\bfx} = \bfU_M \bfc
\end{equation}
where the coefficient vector $\bfc$ is given by
\begin{equation}\label{eq:forward}
\textstyle
    \bfc = \bfU_M^t \bfx.
\end{equation}
Thus, we represent the $256$-dimensional TF $\bfx$ with the $M$-dimensional vector $\bfc$ in \eqref{eq:forward}. As a result, we need to predict only $M$ parameters for the intensity enhancement. We set $M=10$ based on an ablation study in Section ~\ref{sec:exp}.

\begin{figure}[t]
\centering 
\includegraphics[width=1\linewidth]{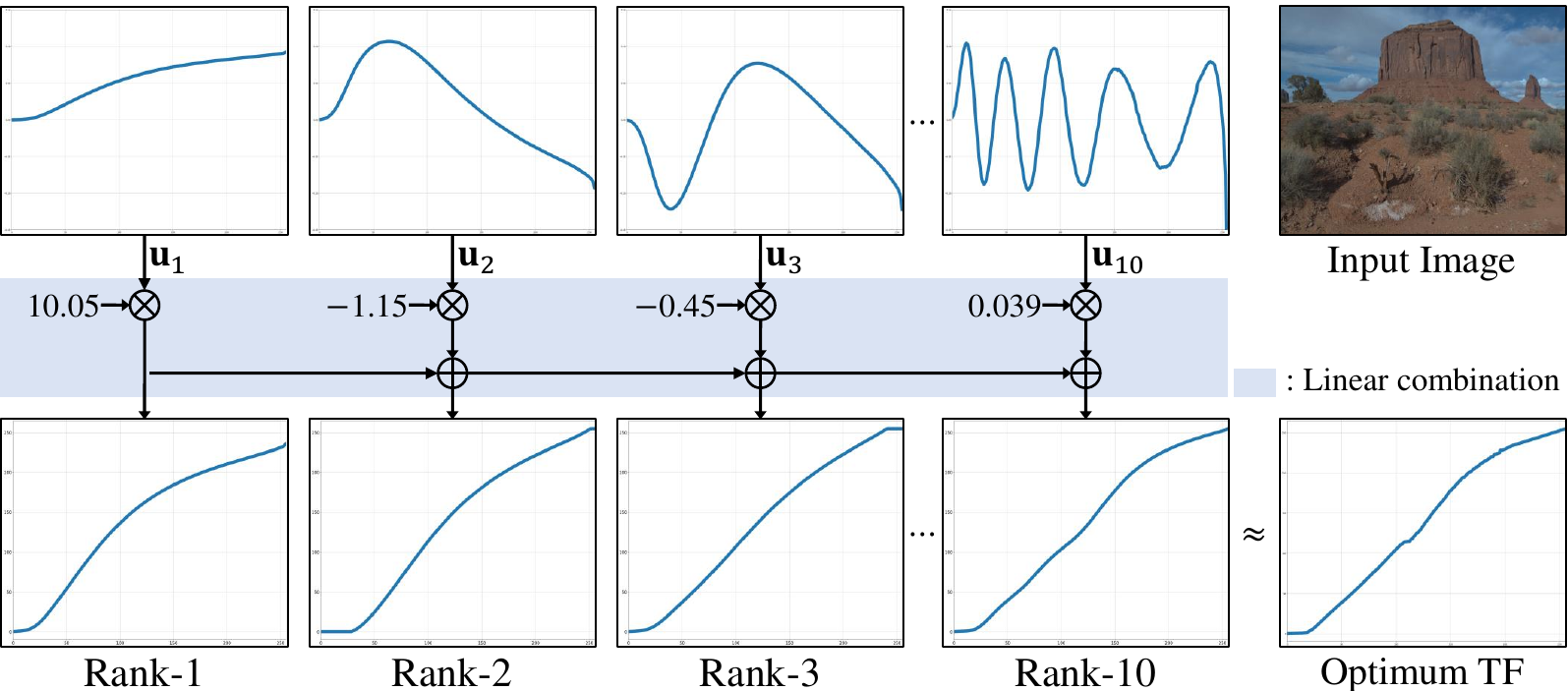}
\caption{Illustration of the eigenTF representation. As more eigenTFs $\bfu_i$ are used for the linear combination, the reconstructed TF gets more faithful to the optimum TF. }
\label{fig:eigenTF_examples}
\end{figure}

\begin{figure*}[t]
\centering 
\includegraphics[width=1\linewidth]{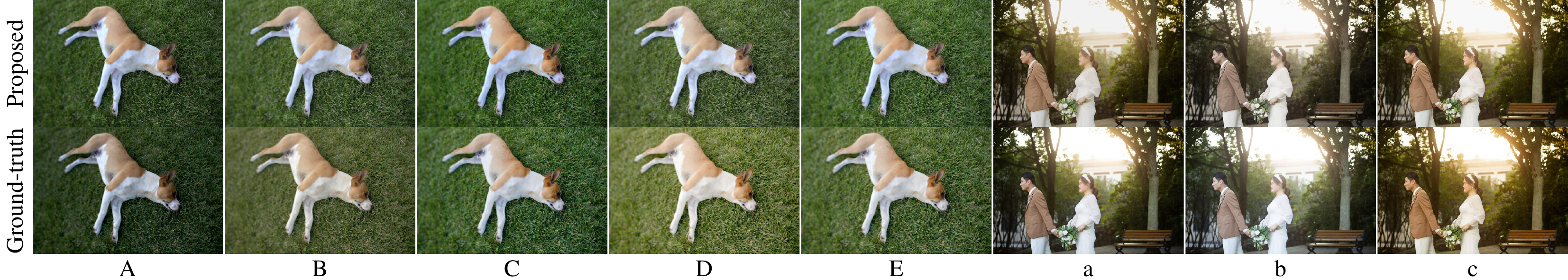}
\caption{Oneta enhancement results for the retouching task: the left five images are for experts A, B, C, D, and E in MIT-Adobe FiveK \cite{bychkovsky2011learning}, and the right three are for experts a, b, and c in PPR10K \cite{liang2021ppr10k}. } 
\label{fig:qualitative_results1}
\end{figure*}

\begin{figure*}[h]
\vspace*{0.32cm} 
\centering 
\includegraphics[width=1\linewidth]{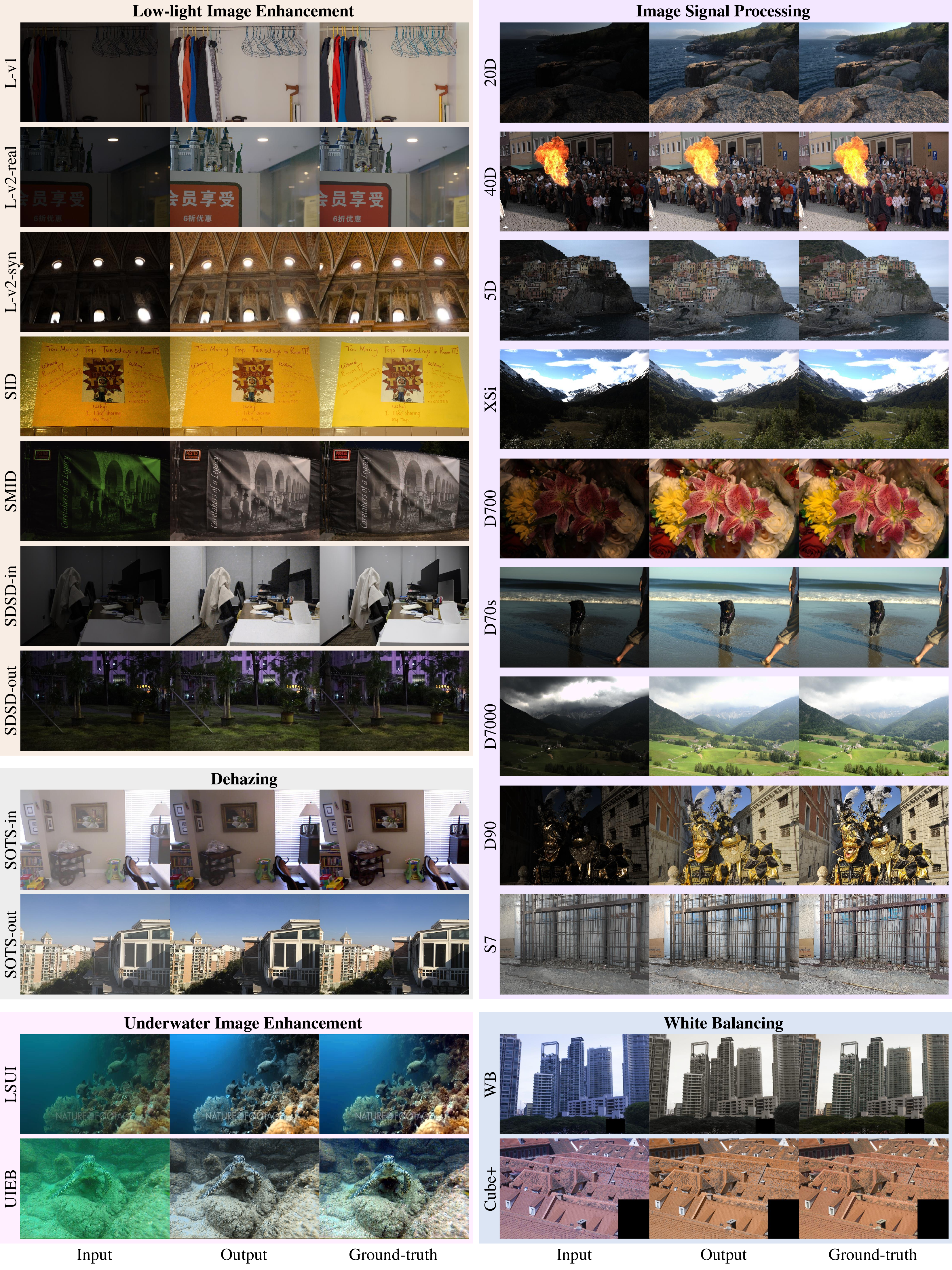}
\caption{Oneta enhancement results for the LLIE, dehazing, UIE, ISP, and WB tasks.} 
\label{fig:qualitative_results2}
\end{figure*}

Figure~\ref{fig:eigenTF_examples} illustrates the low-rank approximation of an optimum TF using the first ten eigenTFs. The first eigenTF $\bfu_1$ is an overall increasing curve, while the others are used to reconstruct the details in the optimum TF. We see that the rank-10 approximation is quite faithful to the optimum TF. 

\subsection{Oneta Network}

As shown in Figure~\ref{fig:overview}, the proposed Oneta consists of Y-Net and C-Net, both based on ViT \cite{dosovitskiy2020image}. Y-Net predicts the coefficient vector $\bfc$ in \eqref{eq:forward} to estimate a TF, while C-Net estimates the CCM in \eqref{eq:CCM}.

\vspace*{0.1cm}
\noindent\textbf{ViT backbone:} We adopt the ViT \cite{dosovitskiy2020image} pretrained on ImageNet \cite{deng2009imagenet} as the backbone of Y-Net and C-Net. An image is split into patches, and the flattened patches are linearly transformed into patch embeddings. Position embeddings are then added to the patch embeddings. Then, the resulting embedding vectors are input to the transformer encoder. The encoder architecture follows  \cite{waswani2017attention}, which consists of layers of multi-head self-attention and MLP blocks with layer normalization and residual connections.

\vspace*{0.1cm}
\noindent\textbf{Style tokens:} In \cite{dosovitskiy2020image}, a single extra token is prepended to a sequence of embedding vectors, and it is used for the classification task. In this paper, we employ as many learnable tokens as the number of styles that Oneta supports. These style tokens are denoted by $\bfs_1, \cdots, \bfs_K$, where $K$ is the number of styles. One of these tokens is selected through a switch depending on the desired style. Each style token $\bfs_i$ is initialized with the pre-trained extra token in ViT \cite{dosovitskiy2020image}. However, as Oneta is trained, each $\bfs_i$ learns to better represent its own style. 

\vspace*{0.1cm}
\noindent\textbf{Y-Net:} First, Y-Net attends the style token $\bfs_i$ to the embedding vectors from $\mathcal{I}_\mathrm{in}$. As a result, it yields the encoded style token $\bfs_i'$, which is fed into a series of three fully connected layers for regressing the $M$ coefficients in $\bfc$ in \eqref{eq:forward}. Then, the enhancement model reconstructs the TF via \eqref{eq:backward} and uses it to transform $\mathcal{I}_\mathrm{in}$ into $\mathcal{I}_\mathrm{mid}$.

\vspace*{0.1cm}
\noindent\textbf{C-Net:}
Next, C-Net further attends $\bfs_i'$ to the embedding vectors from  $\mathcal{I}_\mathrm{mid}$, yielding $\bfs_i''$. It then processes $\bfs_i''$ through three fully connected layers to regress the 6 parameters for the CCM in \eqref{eq:CCM}. The CCM is used to transform $\mathcal{I}_\mathrm{mid}$ into the final output image $\mathcal{I}_\mathrm{out}$.

\vspace*{0.1cm}
\noindent\textbf{Loss functions:} We train Y-Net to minimize the $\ell_2$ loss between the predicted coefficient vector ${\bfc}$ and the optimum $\bfc^*$, obtained by \eqref{eq:QP_TF} and \eqref{eq:forward}, and the $\ell_2$ loss between the intensity channels of the predicted intermediate image and the GT.  Also, we train C-Net to minimize the $\ell_2$ loss between the predicted and GT images. Hence, the total loss is defined as
\begin{equation}\label{eq:loss}
\textstyle
    \mathcal{L} =  \lambda  \|  \bfc-\bfc^*  \|^2 + \|  \mathbf{y}_\mathrm{mid} -\mathbf{y}_\mathrm{GT} \|^2 + \|  \mathcal{I}_\mathrm{out} - \mathcal{I}_\mathrm{GT} \|^2.
\end{equation}
The $\ell_2$ loss between the predicted CCM $\bm \kappa$ and the optimum $\bm{\kappa}^*$ is not included because $\bm{\kappa}^*$ assumes the optimum intermediate image $\mathcal{I}_\mathrm{mid}^*$, which is unavailable during training. 

\section{Experiments}
\label{sec:exp}

\subsection{Experimental Settings}

\noindent\textbf{Datasets}: We use 30 datasets (or styles) for six image enhancement tasks. 
\begin{itemize}
\itemsep1mm
\item Retouching: 5 retouching results of experts A, B, C, D, and E in MIT-Adobe FiveK \cite{bychkovsky2011learning} and 3 retouching results of experts a, b, and c in PPR10K \cite{liang2021ppr10k}. 
\item ISP: MIT-Adobe FiveK \cite{bychkovsky2011learning} for 4 Canon EOS cameras and 2 Nikon cameras, RAISE \cite{dang2015raise} for 2 Nikon cameras, and Samsung Galaxy S7 \cite{schwartz2018deepisp}.
\item LLIE: LOL-v1 \cite{wei2018deep}, LOL-v2-real \cite{yang2021sparse}, LOL-v2-syn \cite{yang2021sparse}, SID \cite{chen2019seeing}, SMID \cite{chen2018learning}, SDSD-in \cite{wang2021seeing}, and SDSD-out \cite{wang2021seeing}.
\item Dehazing: RESIDE-ITS, OTS \cite{li2018benchmarking} for training and SOTS-indoor, outdoor \cite{li2018benchmarking} for testing.
\item UIE: LSUI \cite{peng2023u} and UIEB \cite{li2019underwater}.
\item WB:  WB \cite{afifi2019color} and Cube+ \cite{banic2017unsupervised}.
\end{itemize}
LLIE images are pre-processed with the BT.709 gamma correction \cite{bt2002parameter} and the bilateral filter \cite{tomasi1998bilateral} to reduce noise, which is not removable by point operators. Also, ISP images are pre-processed with the same gamma correction. More details on the datasets are available in the supplement.

\vspace*{0.1cm}
\noindent\textbf{Implementation details}: We initialize Y-Net and C-Net using ViT \cite{dosovitskiy2020image} and the 30 style tokens identically using the extra token in \cite{dosovitskiy2020image}. We use the AdamW optimizer \cite{loshchilov2017decoupled} with an initial learning rate $10^{-4}$, a weight decay $10^{-4}$, $\beta_1 = 0.9$, and $\beta_2 = 0.999$. We halve the learning rate after 30 and 60 epochs, respectively, and train Oneta with a batch size of 16 for 85 epochs in total. Training images are resized to $224 \times 224$, and augmented by random rotation and vertical flipping, while the test images are enhanced at the original sizes. We set $\lambda=0.1$ in \eqref{eq:loss}.

\begin{table*}[t]
\centering
\vspace*{-0.3cm} 
\renewcommand{\arraystretch}{1.0}
\fontsize{7.5pt}{9pt}\selectfont
\caption{Comparison of the performances on various image enhancement tasks.}
\vspace*{-0.2cm}

\begin{minipage}{1.0\linewidth}  
    \centering
    \fontsize{7.0pt}{8pt}\selectfont
    \resizebox{\textwidth}{!}{
    \begin{tabularx}{\textwidth}{l>{\centering\arraybackslash}X>{\centering\arraybackslash}X>{\centering\arraybackslash}X>{\centering\arraybackslash}X>{\centering\arraybackslash}X>{\centering\arraybackslash}X>{\centering\arraybackslash}X>{\centering\arraybackslash}X}
    \toprule
    {\textbf{Retouching}} & A & B & C & D & E & a & b & c \\
    \midrule
    SepLUT \cite{yang2022seplut} & 22.71 & 26.15 & 25.47 & 23.09 & 24.42 & 26.28 & 25.23 & 25.59 \\
    BGridLUT \cite{kim2024image} & 22.65 & 26.24 & 25.66 & 23.25 & 24.34 & 26.45 & 25.48 & 25.72 \\
    \midrule
    Proposed & 22.27 & 25.37 & 23.91 & 22.51 & 23.40 & 21.80 & 21.57 & 22.66 \\
    Upper bound & 32.10 & 32.33 & 30.59 & 31.50 & 29.43 & 32.40 & 33.48 & 32.91 \\
    \bottomrule
    \end{tabularx}}
\end{minipage}

\vspace*{0.1cm} 

\begin{minipage}{1.0\linewidth}  
    \centering
    \fontsize{7.0pt}{8pt}\selectfont
    \resizebox{\textwidth}{!}{
    \begin{tabularx}{\textwidth}{l>{\centering\arraybackslash}X>{\centering\arraybackslash}X>{\centering\arraybackslash}X>{\centering\arraybackslash}X>{\centering\arraybackslash}X>{\centering\arraybackslash}X>{\centering\arraybackslash}X>{\centering\arraybackslash}X>{\centering\arraybackslash}X}

    \toprule
    {\textbf{ISP}} & 20D & 40D & 5D & XSi & D700 & D70s & D7000 & D90 & S7\\
    \midrule
    InvISP \cite{xing2021invertible} & 31.13 & 28.48 & 33.61 & 29.64 & 37.47 & 28.90 & 30.20 & 28.89 & 28.96 \\
    ParamISP \cite{kim2023paramisp} & - & - & - & - & - & - & 34.14 & 30.83 & 29.02 \\
    \midrule
    Proposed & 29.21 & 28.53 & 29.31 & 29.38 & 29.69 & 28.89 & 23.86 & 22.28 & 24.37 \\
    Upper bound & 32.19 & 30.85 & 31.58 & 32.48 & 32.68 & 31.66 & 29.12 & 25.70 & 27.03 \\
    \bottomrule
    \end{tabularx}}
\end{minipage}

\vspace*{0.1cm} 

\begin{minipage}{1.0\linewidth}  
    \centering
    \fontsize{7.0pt}{8pt}\selectfont
    \resizebox{\textwidth}{!}{
    \begin{tabularx}{\textwidth}{l>{\centering\arraybackslash}X>{\centering\arraybackslash}X>{\centering\arraybackslash}X>{\centering\arraybackslash}X>{\centering\arraybackslash}X>{\centering\arraybackslash}X>{\centering\arraybackslash}X}
    \toprule
    {\textbf{LLIE}} & L-v1 & L-v2-real & L-v2-syn & SID & SMID & SDSD-in & SDSD-out \\
    \midrule
    SNR-Net \cite{xu2022snr} & 24.61 & 21.48 & 24.14 & 22.87 & 28.49 & 29.44 & 28.66 \\
    Retinexformer \cite{cai2023retinexformer} & 25.16 & 22.80 & 25.67 & 24.44 & 29.15 & 29.77 & 29.84 \\
    \midrule
    Proposed & 24.89 & 27.91 & 22.15 & 18.23 & 25.56 & 27.06 & 28.19 \\
    Upper bound & 27.61 & 30.08 & 30.00 & 20.40 & 28.36 & 30.27 & 31.77 \\
    \bottomrule
    \end{tabularx}}
\end{minipage}

\vspace*{0.1cm} 

\hspace{0.01em}
\begin{minipage}{0.320\linewidth}
    \centering
    \fontsize{7.0pt}{8pt}\selectfont
    \begin{tabularx}{\linewidth}{l>{\centering\arraybackslash}X>{\centering\arraybackslash}X}
        \toprule
        {\textbf{Dehazing}} & SOTS-in & SOTS-out  \\
        \midrule
         PMNet\cite{ye2022perceiving} & 38.41 & 34.74 \\
         C$^2$PNet \cite{zheng2023curricular} & 42.56 & 36.68 \\
        \midrule
        Proposed & 20.66 & 28.51 \\
        Upper bound & 22.35 & 29.92 \\
        \bottomrule
    \end{tabularx}
\end{minipage}
\hspace{0.6em}
\begin{minipage}{0.320\linewidth}
    \centering
    \fontsize{7.0pt}{8pt}\selectfont
    \begin{tabularx}{\linewidth}{l>{\centering\arraybackslash}X>{\centering\arraybackslash}X}
        \toprule
         {\textbf{UIE}}  & LSUI & UIEB \\
        \midrule
         Ucolor \cite{li2021underwater} & 22.91 & 20.78 \\
         U-shape \cite{peng2023u}  & 24.16 & 22.91 \\
        \midrule
        Proposed & 24.90 & 22.15(19.87) \\
        Upper bound & 27.36 & 27.09 \\
        \bottomrule
    \end{tabularx}
\end{minipage}
\hspace{0.6em}
\begin{minipage}{0.320\linewidth}
    \centering
    \fontsize{7.0pt}{8pt}\selectfont
    \begin{tabularx}{\linewidth}{l>{\centering\arraybackslash}X>{\centering\arraybackslash}X}
        \toprule
        {\textbf{WB}} & WB & Cube+  \\
        \midrule
         DeepWB \cite{afifi2020deep} & 28.96 & 29.04 \\
         STAR \cite{zhang2021star} & 29.10 & 29.33  \\
        \midrule
        Proposed & 27.10 & 29.54(28.92) \\
        Upper bound & 29.10 & 30.32 \\
        \bottomrule
    \end{tabularx}
\end{minipage}
\label{table:comparison}
\end{table*}

\subsection{Multi-Style Image Enhancement}
Figures~\ref{fig:qualitative_results1} and \ref{fig:qualitative_results2} show Oneta enhancement examples, one for each style. In Figure~\ref{fig:qualitative_results1}, for the dog image, the output images for styles A, C, and E are more faithful to the GT than those for styles B and D are. However, the retouched images match the different GT styles decently. In Figure~\ref{fig:qualitative_results2}, Oneta performs LLIE, dehazing, UIE, ISP, and WB. SID is a challenging dataset for LLIE. In this SID example, Oneta provides different color tones from the GT, but it still achieves the objective of LLIE, improving the overall brightness and exhibiting the details of the center poster clearly. For the two UIE examples, Oneta also yields different tones from the GT. But, for the other cases, we see that Oneta output is quite faithful to the corresponding GT. It is worth reiterating that all these 30 styles are supported by the single Oneta network. 

Figure~\ref{fig:use_cases} presents several use cases of Oneta. Figure~\ref{fig:use_cases}(a) retouches a single image in the 8 different styles, and Figure~\ref{fig:use_cases}(b) improves a low-light image in the 7 different styles. Both results confirm the subjective nature of image enhancement \cite{DIP_BOOK_Jain}; it is hard to tell the best results from those multiple candidates. Oneta can support users' subjective preferences by allowing them to experiment with different style tokens and choose the most preferred results.

Figure~\ref{fig:use_cases}(c) tests the interpolation of style tokens. The top row mixes the two style tokens for retouching experts B and C, while the bottom row does the tokens for experts E and C. We see that the mixed tokens provide retouched images in the intermediate styles. Thus, Oneta can utilize not only the learned style tokens but also many interpolated ones. 

Figure~\ref{fig:use_cases}(d) illustrates that a user can perform the enhancement by employing multiple style tokens in series. These use cases show that users can utilize the single Oneta network to enhance images in various manners according to their preferences. 

The supplemental document provides many more enhancement results of Oneta. 

\subsection{Quantitative Results}
Table~\ref{table:comparison} summarizes the PSNR performance of the proposed Oneta on each dataset. Although PSNR is not the best metric for subjective enhancement tasks, it is adopted for its objectivity and ease of computations. For comparison, for each task, we select recent high-performing algorithms. These algorithms are designed for specific tasks or datasets only. For example, most retouching algorithms consider only the style of expert C in MIT-Adobe FiveK. Thus, we train SepLUT \cite{yang2022seplut} and BGridLUT \cite{kim2024image} on experts A, B, D, and E, respectively. Also, we train InvISP \cite{xing2021invertible} on 20D, 40D, XSi, and D70s using the available source code. The following observations can be made from Table~\ref{table:comparison}: 
\vspace*{1mm}
\begin{itemize}
\itemsep1mm
\item The table also lists the PSNR upper bounds of the proposed two-step enhancement model on the test datasets, which are relatively high --- higher than the results of the high-performing existing algorithms --- for most datasets. 

\begin{figure}[!h]
\centering 
\vspace*{0.1cm}
\includegraphics[width=1\linewidth]{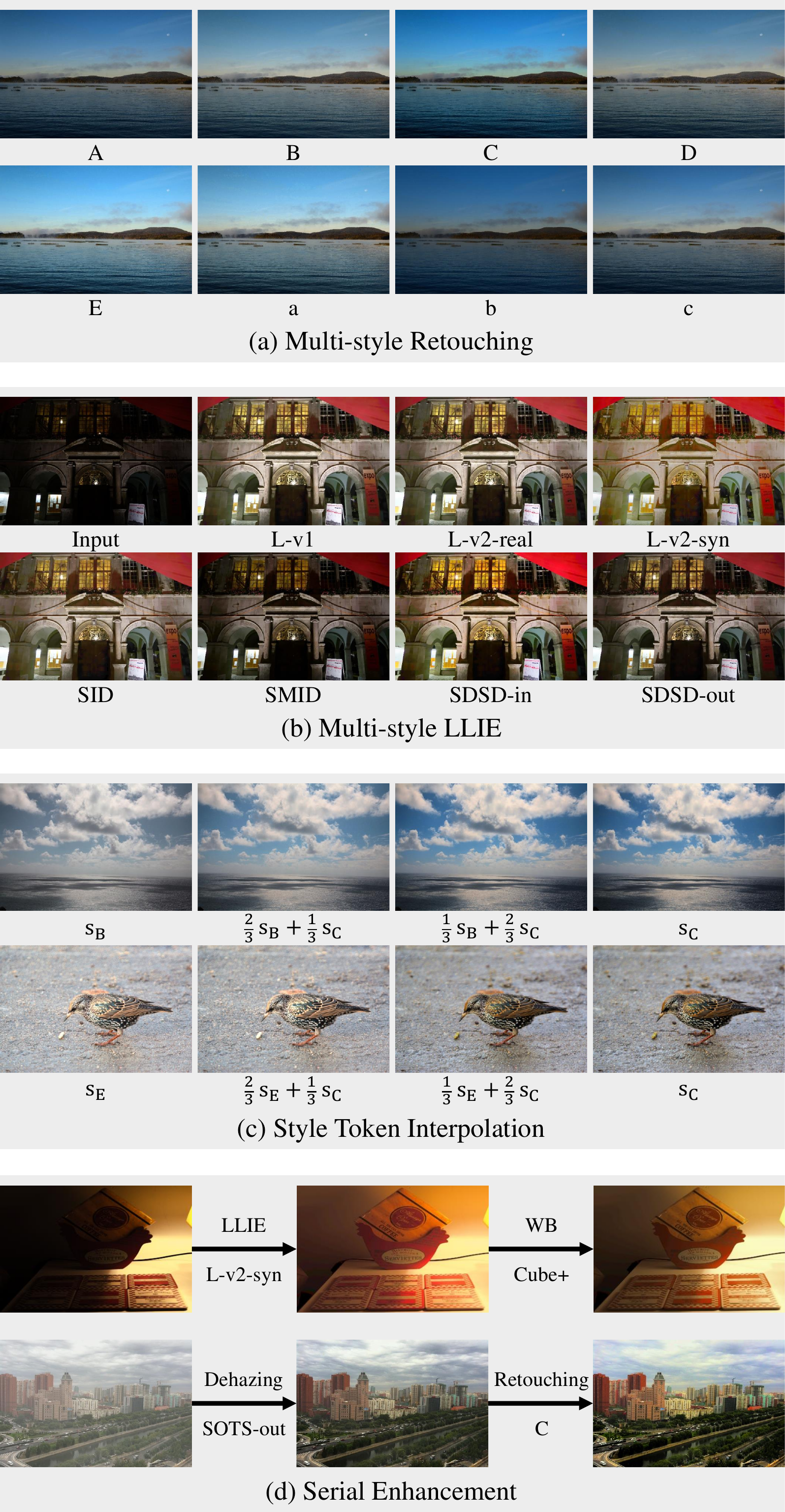}
\caption{Several use cases of Oneta.} 
\label{fig:use_cases}
\vspace*{-1.0cm}
\end{figure}

\item Except retouching (a, b, c), ISP (D700, D7000, D90, S7), LLIE (SID), dehazing (SOTS-in/out), Oneta provides comparable PSNRs to the existing algorithms on the remaining 20 datasets. This is commendable because the existing algorithms are optimized for their specific tasks or datasets, whereas Oneta performs multi-style enhancement. 

\begin{figure}[t]
\centering 
\includegraphics[width=1\linewidth]{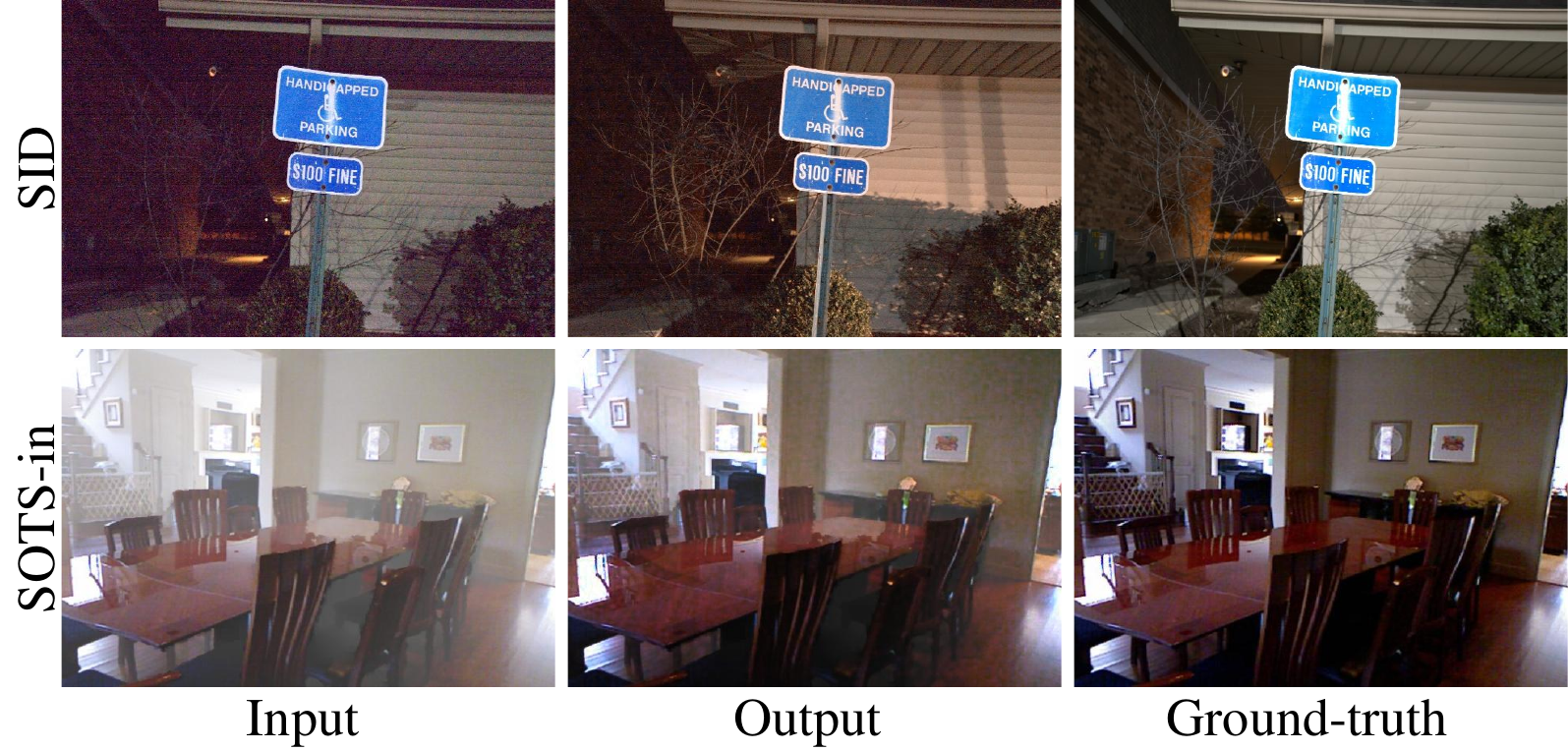}
\caption{In the top example, Oneta fails to improve the brightness sufficiently and reconstruct the tones in the GT. In the bottom example, Oneta removes haze around the nearby table and chairs but fails to do so at the far side of the room.}
\label{fig:failure case}
\end{figure}

\item Despite relatively low PSNRs on retouching (a, b, c) and ISP (D700, D7000, D90, S7), Oneta yields reasonably good results on them; it retouches an image or processes a raw image in different styles well, as shown in Figure~\ref{fig:qualitative_results1}, Figure~\ref{fig:qualitative_results2}, and the supplement. On the other hand, it appears that the proposed two-step global enhancement is not sufficient for LLIE (SID) and dehazing (SOTS-in/out). Figure~\ref{fig:failure case} shows failure cases. For these datasets, additional local enhancement would be helpful. 

\item Oneta outperforms InvISP \cite{xing2021invertible} and U-shape \cite{peng2023u} on the ISP (40D) and UIE (LSUI) datasets, respectively. In UIE, the existing algorithms trained on LSUI are tested on UIEB. Similarly, in WB, the algorithms trained on WB are tested on Cube+. Hence, for a fair comparison, we measure the performance of Oneta on UIEB and Cube+ using the LSUI and WB style tokens, respectively, and report these PSNRs within the parentheses in Table~\ref{table:comparison}. Oneta provides decent PSNRs in these cases. 

\item Oneta outperforms the existing algorithms on LLIE (LOL-v2-real) by a significant margin of more than 5.1dB. If Oneta is trained only on LOL-v2-real, the PSNR is reduced to 21.58dB. This means that Oneta benefits from learning to enhance diverse images and thus yields reliable performance on LOL-v2-real containing only 689 training images. 
\end{itemize}

\subsection{Analysis}

\begin{figure}[t]
\centering 
\includegraphics[width=1\linewidth]{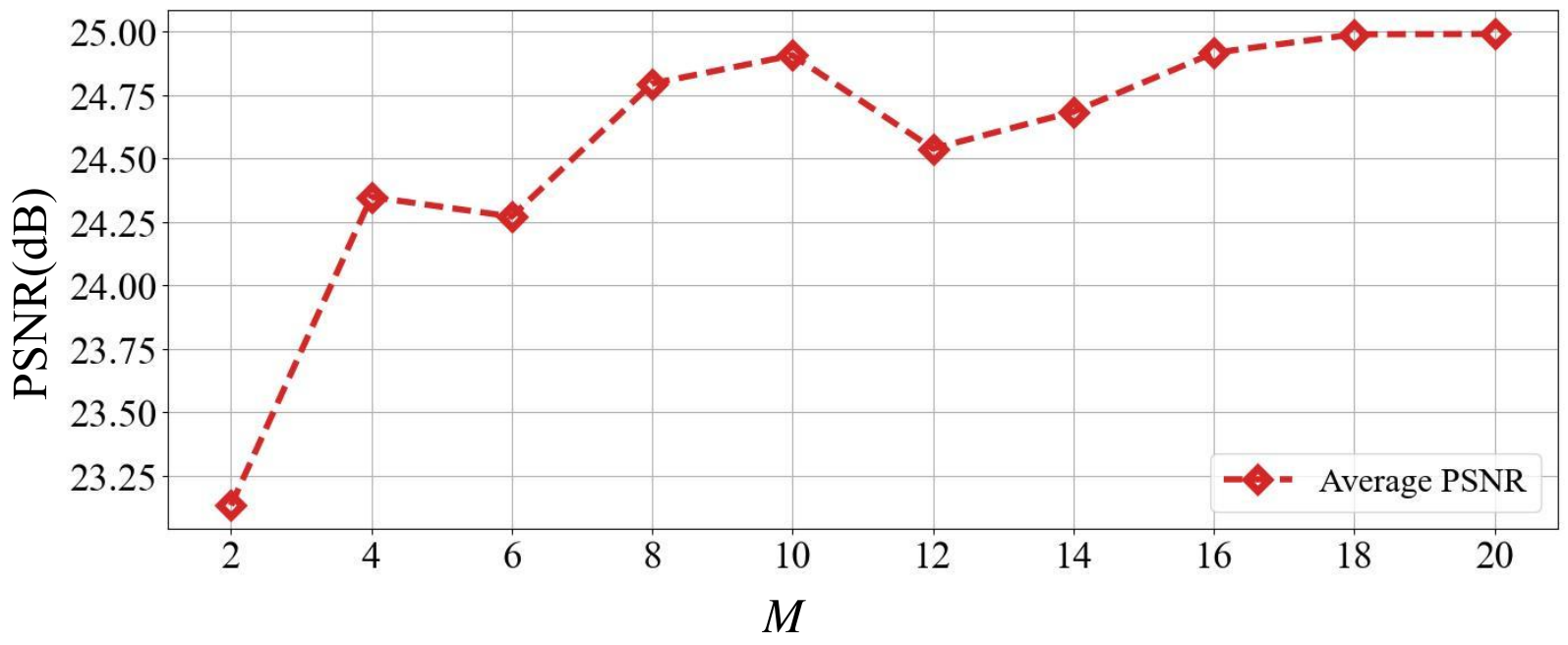}
\caption{Ablation study on the number $M$ of eigenTFs.} 
\label{fig:m_ablation}
\end{figure}

\noindent\textbf{Performance according to $M$}: Figure~\ref{fig:m_ablation} shows an ablation study about the number $M$ of eigenTFs in \eqref{eq:x_approx}, on  a simplified dataset described in the supplement. A bigger $M$ provides better performance in general, but the average PSNR saturates when $M > 10$. Thus, we set $M=10$. 

\vspace*{0.1cm}
\noindent\textbf{Uni-style training}: For comparison, we train Oneta on each dataset separately. This uni-style training is not only inefficient but also underperforms on most datasets, as detailed in the supplement. In other words, Oneta provides more reliable and better enhancement results by learning from diverse tasks simultaneously. 
 
\begin{figure}[t]
\centering 
\includegraphics[width=1\linewidth]{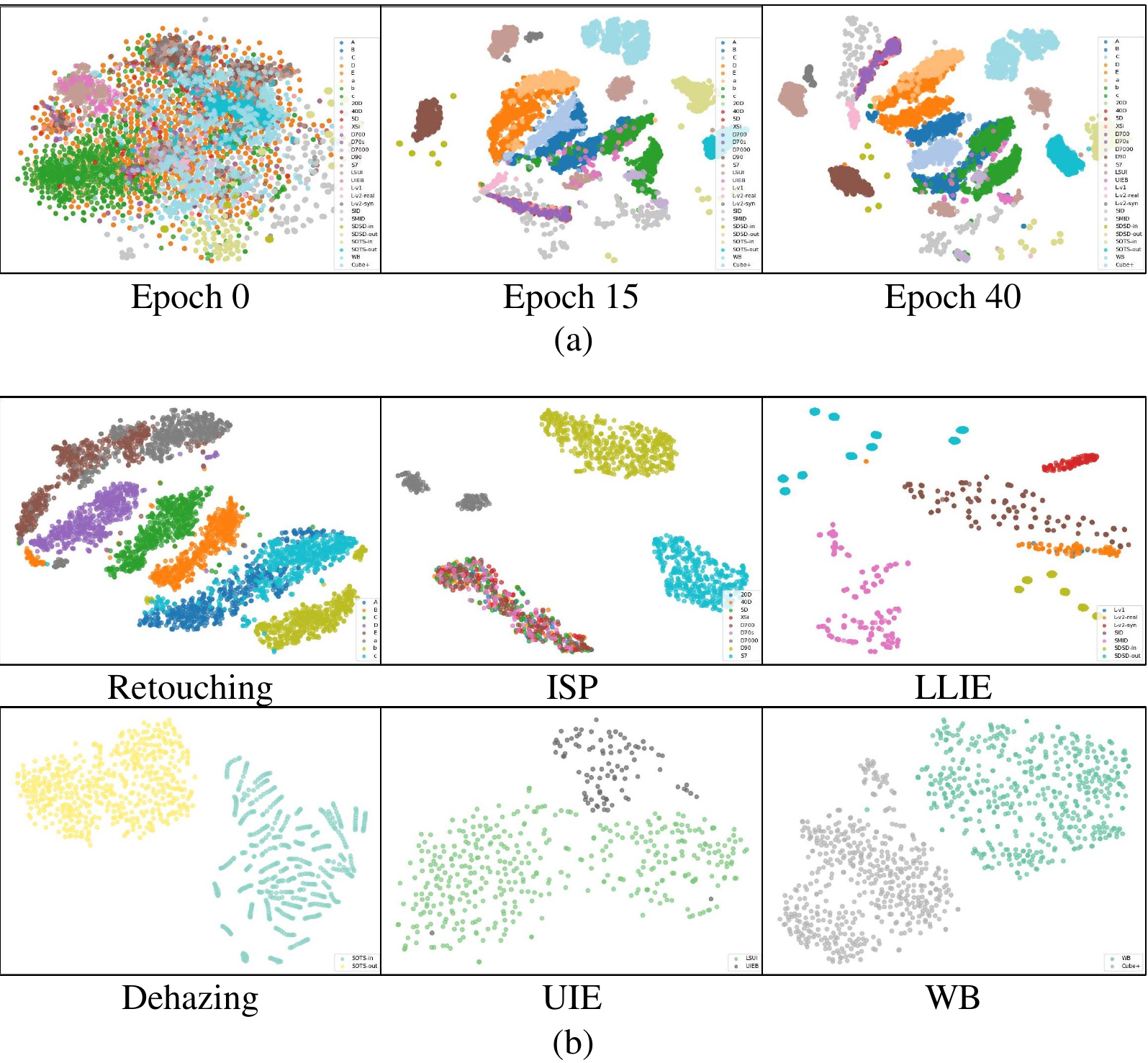}
\caption{t-SNE visualization: (a) clustering of style tokens  $\bfs_i'$ during the training and (b) style tokens $\bfs_i'$ for each task.} 
\label{fig:tsne}
\end{figure}

\vspace*{0.1cm}
\noindent\textbf{Style token visualization}: Figure~\ref{fig:tsne}(a) visualizes how the attended style tokens $\bfs_i'$ of the 30 datasets form clusters, as the train goes on, using t-SNE \cite{van2008visualizing}. They are eventually clustered according to their styles (or datasets). Figure~\ref{fig:tsne}(b) renders the style tokens $\bfs_i'$ for each of the six tasks separately. More t-SNE visualizations of $\bfs_i$, $\bfs_i'$, and $\bfs_i''$ are  in the supplement.

\section{Conclusions}
The first algorithm Oneta for multi-style image enhancement was proposed in this paper. Oneta can perform various enhancement tasks using a single network by switching a style token. To this end, we developed a simple two-step enhancement model: intensity enhancement through a TF and color correction through a CCM. Moreover, we introduced eigenTFs to represent TFs compactly. Extensive experiments showed that Oneta can cope with six image enhancement tasks of retouching, ISP, LLIE, dehazing, UIE, and WB across 30 datasets. Specifically, on a majority of the datasets, the single Oneta network provides comparable enhancement results to the existing algorithms optimized for specific tasks or datasets.

{
    \small
    \bibliographystyle{ieeenat_fullname}
    \bibliography{main}
}



\appendix

{\onecolumn
\centering
\Large
\textbf{\thetitle}\\
\vspace{0.5em}Supplementary Material\\
}

\section{Datasets}

\begin{table*}[b]\centering
\vspace*{-0.1cm}
    \renewcommand{\arraystretch}{0.8}
    \caption
    {
        Datasets used in the experiments. For each dataset, we specify the short name assigned for convenience, the number of training images, the number of test images, and  whether it is used for the ablation study in Figure~\ref{fig:m_ablation}.
    }
    \fontsize{6.5pt}{\baselineskip}\selectfont
    \vspace*{-0.15cm}
    \resizebox{0.95\linewidth}{!}{
    \begin{tabular}[t]{+L{1.2cm}^L{2.5cm}^L{1.2cm}^C{1.0cm}^C{1.0cm}^C{1.0cm}}
    \toprule
    Task & Dataset & Short name & \# training & \# test & Ablation \\
    \midrule
         \multirow{8}{*}{Retouching} 
         & FiveK-expert A  \cite{bychkovsky2011learning} & A & 4502 & 498 & $\times$ \\
         & FiveK-expert B  \cite{bychkovsky2011learning} & B & 4502 & 498 & $\times$ \\
         & FiveK-expert C  \cite{bychkovsky2011learning} & C & 4502 & 498 & $\ocircle$ \\
         & FiveK-expert D  \cite{bychkovsky2011learning} & D & 4502 & 498 & $\times$\\
         & FiveK-expert E  \cite{bychkovsky2011learning} & E & 4502 & 498 & $\times$\\
         & PPR10K-expert a \cite{liang2021ppr10k} & a & 4500 & 2286 & $\times$\\
         & PPR10K-expert b \cite{liang2021ppr10k} & b & 4500 & 2286 & $\ocircle$\\
         & PPR10K-expert c \cite{liang2021ppr10k} & c & 4500 & 2286 & $\times$\\
     \midrule
         \multirow{9}{*}{ISP} 
         & FiveK-CanonEOS 20D   \cite{bychkovsky2011learning}& 20D & 311 & 54 & $\times$ \\
         & FiveK-Canon EOS 40D   \cite{bychkovsky2011learning}& 40D & 267 & 47 & $\times$ \\
         & FiveK-Canon EOS 5D   \cite{bychkovsky2011learning}& 5D & 661 & 116 & $\ocircle$ \\
         & FiveK-Canon EOS XSi   \cite{bychkovsky2011learning} & XSi & 533 & 94 & $\times$ \\
         & FiveK-Nikon D700  \cite{bychkovsky2011learning} & D700 & 502 & 88 & $\ocircle$ \\
         & FiveK-Nikon D70s  \cite{bychkovsky2011learning} & D70s & 388 & 68 & $\times$ \\
         & RAISE-Nikon D7000 \cite{dang2015raise} & D7000 & 4797 & 1000 & $\times$ \\
         & RAISE-Nikon D90   \cite{dang2015raise} & D90 & 1875 & 400 & $\times$ \\
         & Samsung Galaxy S7 \cite{schwartz2018deepisp}  & S7 & 70 & 150 & $\times$ \\
     \midrule
         \multirow{7}{*}{LLIE}     
         & LOL-v1 \cite{wei2018deep} & L-v1 & 485 & 15 & $\times$ \\
         & LOL-v2-real \cite{yang2021sparse} & L-v2-real & 689 & 100 & $\ocircle$ \\
         & LOL-v2-syn \cite{yang2021sparse} & L-v2-syn & 900 & 100 & $\times$ \\
         & SID \cite{chen2019seeing} & SID & 2157 & 540 & $\times$ \\
         & SMID \cite{chen2018learning} & SMID & 15763 & 5046 & $\times$ \\
         & SDSD-indoor \cite{wang2021seeing} & SDSD-in & 1786 & 177 & $\ocircle$ \\
         & SDSD-outdoor \cite{wang2021seeing} & SDSD-out & 2850 & 300 & $\times$ \\
     \midrule
         \multirow{2}{*}{Dehazing}
         & RESIDE-ITS \cite{li2018benchmarking} & SOTS-in & 13990 & 500 & $\times$ \\
         & RESIDE-OTS \cite{li2018benchmarking} & SOTS-out & 31374 & 500 & $\ocircle$ \\
     \midrule
         \multirow{2}{*}{UIE}
         & LSUI \cite{peng2023u} & LSUI & 3879 & 400 & $\times$ \\
         & UIEB \cite{li2019underwater} & UIEB & 800 & 90 & $\ocircle$ \\
     \midrule
         \multirow{2}{*}{WB}
         & WB \cite{afifi2019color} & WB & 12000 & 21046 & $\ocircle$ \\
         & Cube+ \cite{banic2017unsupervised} & Cube+ & 9000 & 1242 & $\times$ \\        
    \bottomrule
    \end{tabular}}
    \label{table:dataset details}
\end{table*}

\noindent We use 30 datasets (or styles) for six image enhancement tasks: retouching, ISP, UIE, LLIE, dehazing, and WB. For seven ISP datasets (20D, 40D, XSi, D70s, D7000, D90, S7), training images are randomly selected, as they are not specified by the comparison algorithms.

\vspace*{0.1cm}
\noindent\textbf{Retouching}:
Retouching is a task to adjust the brightness and color of an image in various styles. We use two popular datasets for retouching: MIT-Adobe FiveK  \cite{bychkovsky2011learning} and PPR10K \cite{liang2021ppr10k}. FiveK contains manually retouched images by five experts A/B/C/D/E, and PPR10K includes results by three experts a/b/c.

\vspace*{0.1cm}
\noindent\textbf{ISP}: It is a task to convert a raw image captured by a camera into an sRGB image that closely matches the human visual experience. We collect the raw files of the four Canon cameras (5D, 20D, 40D, XSi) and the two Nikon cameras (D700, D70s) in  MIT-Adobe FiveK \cite{bychkovsky2011learning}. We also gather the raw files of Nikon D7000 and Nikon D90 in the RAISE dataset \cite{dang2015raise} and those of Samsung Galaxy S7 from \cite{schwartz2018deepisp}. Input images are extracted from the NEF or DNG files using \texttt{rawpy} and then pre-processed with the BT.709 gamma correction \cite{bt2002parameter}. GT sRGB images for the six FiveK cameras are processed by \texttt{rawpy}, while GTs for D7000, D90, S7 are the JPEG images processed by the cameras. FiveK images are resized to 480p in the test phase.

\vspace*{0.1cm}
\noindent\textbf{LLIE}: It aims to enhance the brightness and color of images captured in low-light conditions, which often exhibit high noise levels. We use the most popular LOL datasets LOL-v1 \cite{wei2018deep}, LOL-v2-real \cite{yang2021sparse}, and LOL-v2-syn \cite{yang2021sparse}. We also use SID \cite{chen2019seeing}, SMID \cite{chen2018learning}, SDSD-in \cite{wang2021seeing} and SDSD-out \cite{wang2021seeing}. Except for LOL-v2-syn \cite{yang2021sparse}, all datasets contain multiple images of the same scene captured under different exposure conditions. We pre-process input images with the BT.709 gamma correction \cite{bt2002parameter} and the bilateral filter \cite{tomasi1998bilateral} sequentially to reduce noise. 

\vspace*{0.1cm}
\noindent\textbf{Dehazing}: It attempts to remove image distortions caused by haze or fog in the atmosphere. We use RESIDE-ITS \cite{li2018benchmarking} and RESIDE-OTS \cite{li2018benchmarking} for training and SOTS-indoor \cite{li2018benchmarking} and SOTS-outdoor \cite{li2018benchmarking} for testing. We reduce the number of training images by a factor of 1/10 to prevent overfitting on the RESIDE dataset. RESIDE images consist of multiple data samples of the same scene with varying degrees of haze.

\vspace*{0.1cm}
\noindent\textbf{UIE}: It is a task to improve the quality of underwater images by addressing color distortions, low contrast, and blurriness. We use two datasets LSUI \cite{peng2023u} and UIEB \cite{li2019underwater}, which contain real underwater images and the corresponding post-processed GT images. 

\vspace*{0.1cm}
\noindent\textbf{WB}: WB is a task that adjusts the color temperature of images captured under various lighting conditions. We use the WB \cite{afifi2019color} and Cube+ \cite{banic2017unsupervised} datasets. They contain rendered images from different illumination setups such as fluorescent, incandescent, and daylight. For each rendered image, the corresponding white-balanced GT is also included in the dataset.  

\vspace*{1cm}

\section{Optimum CCM}
Let $\bm{\kappa}$ be the 9-dimensional column vector reshaped from the CCM in \eqref{eq:CCM}. Also, let $(r_i, g_i, b_i)$ denote the $i$th pixel in $\mathcal{I}_\mathrm{mid}^*$ in the raster scan order. To find the optimum CCM $\bm{\kappa}^*$ in \eqref{eq:QP_CCM}, we construct a $3N\times 9$ matrix $\mathbf{B}$ from $\mathcal{I}_\mathrm{mid}^*$, given by
\begin{equation} 
\textstyle
\mathbf{B} = 
\begin{pmatrix}
r_1 & g_1 & b_1 & 0 & 0 & 0 & 0 & 0 & 0 \\
0 & 0 & 0 & r_1 & g_1 & b_1 & 0 & 0 & 0 \\
0 & 0 & 0 & 0 & 0 & 0 & r_1 & g_1 & b_1 \\
\vdots  & \vdots & \vdots & \vdots  & \vdots & \vdots & \vdots & \vdots & \vdots \\
r_N & g_N & b_N & 0 & 0 & 0 & 0 & 0 & 0 \\
0 & 0 & 0 & r_N & g_N & b_N & 0 & 0 & 0 \\
0 & 0 & 0 & 0 & 0 & 0 & r_N & g_N & b_N \\
\end{pmatrix}
\end{equation}
where $N$ is the number of pixels in an image. Then, the product $\mathbf{B}\bm{\kappa}$ represents the color-corrected RGB values. Also, we form a vector $\mathbf{z}_\mathrm{GT}$ composed of the $3N$ RGB values in the GT image $\mathcal{I}_\mathrm{GT}$.

\vspace*{1cm}

\section{Experimental Setting for Table~\ref{table:comparison}}
In Table~\ref{table:comparison}, Ucolor \cite{li2021underwater} and U-shape \cite{peng2023u} are trained on the LSUI dataset and tested on both LSUI and UIEB. Also, DeepWB \cite{afifi2020deep} and STAR \cite{zhang2021star} are trained on the WB dataset and tested on both WB and Cube+. Except for these two cases, all existing algorithms in Table~\ref{table:comparison} are trained separately for each dataset. 

\clearpage
\section{Analysis}

\subsection{Performance according to $M$}
For the ablation study in Figure~\ref{fig:m_ablation}, we use nine datasets specified in Table~\ref{table:dataset details}: FiveK-expert C, PPR10K-expert~b, 
Canon EOS 5D, Nikon D700, UIEB, LOL-v2-real, SDSD-in, RESIDE-OTS, and WB. From each of these datasets, we randomly sample 500 training images and 50 test images. 

\subsection{Uni-style training}

For comparison, we train Oneta on each dataset separately. Table \ref{table:ablation2} compares the results of this uni-style training with those of the proposed multi-style training. Note that multi-style training outperforms uni-style training in most cases.

\begin{table*}[h]
\centering
\vspace*{0.3cm} 
\renewcommand{\arraystretch}{1.0}
\fontsize{7.5pt}{9pt}\selectfont
\caption{Performance comparison between uni-style training and multi-style training.}
\vspace*{-0.2cm}

\begin{minipage}{1.0\linewidth}  
    \centering
    \fontsize{7.0pt}{8pt}\selectfont
    \resizebox{\textwidth}{!}{
    \begin{tabularx}{\textwidth}{l>{\centering\arraybackslash}X>{\centering\arraybackslash}X>{\centering\arraybackslash}X>{\centering\arraybackslash}X>{\centering\arraybackslash}X>{\centering\arraybackslash}X>{\centering\arraybackslash}X>{\centering\arraybackslash}X}
    \toprule
    {\textbf{Retouching}} & A & B & C & D & E & a & b & c \\
    \midrule
    Uni-style & 22.22 & 22.25 & 23.96 & 22.60 & 23.49 & 18.53 & 18.78 & 20.64 \\
    Multi-style & 22.27 & 25.37 & 23.91 & 22.51 & 23.40 & 21.80 & 21.57 & 22.66 \\
    \bottomrule
    \end{tabularx}}
\end{minipage}

\vspace*{0.1cm} 

\begin{minipage}{1.0\linewidth}  
    \centering
    \fontsize{7.0pt}{8pt}\selectfont
    \resizebox{\textwidth}{!}{
    \begin{tabularx}{\textwidth}{l>{\centering\arraybackslash}X>{\centering\arraybackslash}X>{\centering\arraybackslash}X>{\centering\arraybackslash}X>{\centering\arraybackslash}X>{\centering\arraybackslash}X>{\centering\arraybackslash}X>{\centering\arraybackslash}X>{\centering\arraybackslash}X}

    \toprule
    {\textbf{ISP}} & 20D & 40D & 5D & XSi & D700 & D70s & D7000 & D90 & S7\\
    \midrule
    Uni-style & 27.75 & 27.02 & 29.22 & 28.88 & 29.31 & 27.99 & 22.10 & 20.00 & 23.65\\
    Multi-style & 29.21 & 28.53 & 29.31 & 29.38 & 29.69 & 28.89 & 23.86 & 22.28 & 24.37 \\
    \bottomrule
    \end{tabularx}}
\end{minipage}

\vspace*{0.1cm} 

\begin{minipage}{1.0\linewidth}  
    \centering
    \fontsize{7.0pt}{8pt}\selectfont
    \resizebox{\textwidth}{!}{
    \begin{tabularx}{\textwidth}{l>{\centering\arraybackslash}X>{\centering\arraybackslash}X>{\centering\arraybackslash}X>{\centering\arraybackslash}X>{\centering\arraybackslash}X>{\centering\arraybackslash}X>{\centering\arraybackslash}X}
    \toprule
    {\textbf{LLIE}} & L-v1 & L-v2-real & L-v2-syn & SID & SMID & SDSD-in & SDSD-out \\
    \midrule
    Uni-style & 22.12 & 21.28 & 22.08 & 18.11 & 24.67 & 28.66 & 27.39 \\
    Multi-style & 24.89 & 27.91 & 22.15 & 18.23 & 25.56 & 27.06 & 28.19 \\
    \bottomrule
    \end{tabularx}}
\end{minipage}

\vspace*{0.1cm} 

\hspace{0.01em}
\begin{minipage}{0.320\linewidth}
    \centering
    \fontsize{7.0pt}{8pt}\selectfont
    \begin{tabularx}{\linewidth}{l>{\centering\arraybackslash}X>{\centering\arraybackslash}X}
        \toprule
        {\textbf{Dehazing}} & SOTS-in & SOTS-out  \\
        \midrule
        Uni-style & 21.31 & 27.95 \\
        Multi-style & 20.66 & 28.51 \\
        \bottomrule
    \end{tabularx}
\end{minipage}
\hspace{0.6em}
\begin{minipage}{0.320\linewidth}
    \centering
    \fontsize{7.0pt}{8pt}\selectfont
    \begin{tabularx}{\linewidth}{l>{\centering\arraybackslash}X>{\centering\arraybackslash}X}
        \toprule
         {\textbf{UIE}}  & LSUI & UIEB \\
        \midrule
        Uni-style & 24.87 & 21.96(19.58) \\
        Multi-style & 24.90 & 22.15(19.87) \\
        \bottomrule
    \end{tabularx}
\end{minipage}
\hspace{0.6em}
\begin{minipage}{0.320\linewidth}
    \centering
    \fontsize{7.0pt}{8pt}\selectfont
    \begin{tabularx}{\linewidth}{l>{\centering\arraybackslash}X>{\centering\arraybackslash}X}
        \toprule
        {\textbf{WB}} & WB & Cube+  \\
        \midrule
        Uni-style & 26.33 & 29.00(25.73) \\
        Multi-style & 27.10 & 29.54(28.92) \\
        \bottomrule
    \end{tabularx}
\end{minipage}
\label{table:ablation2}
\end{table*}

\subsection{Style token visualization}
\begin{figure*}[h]
\centering 
\includegraphics[width=1\linewidth]{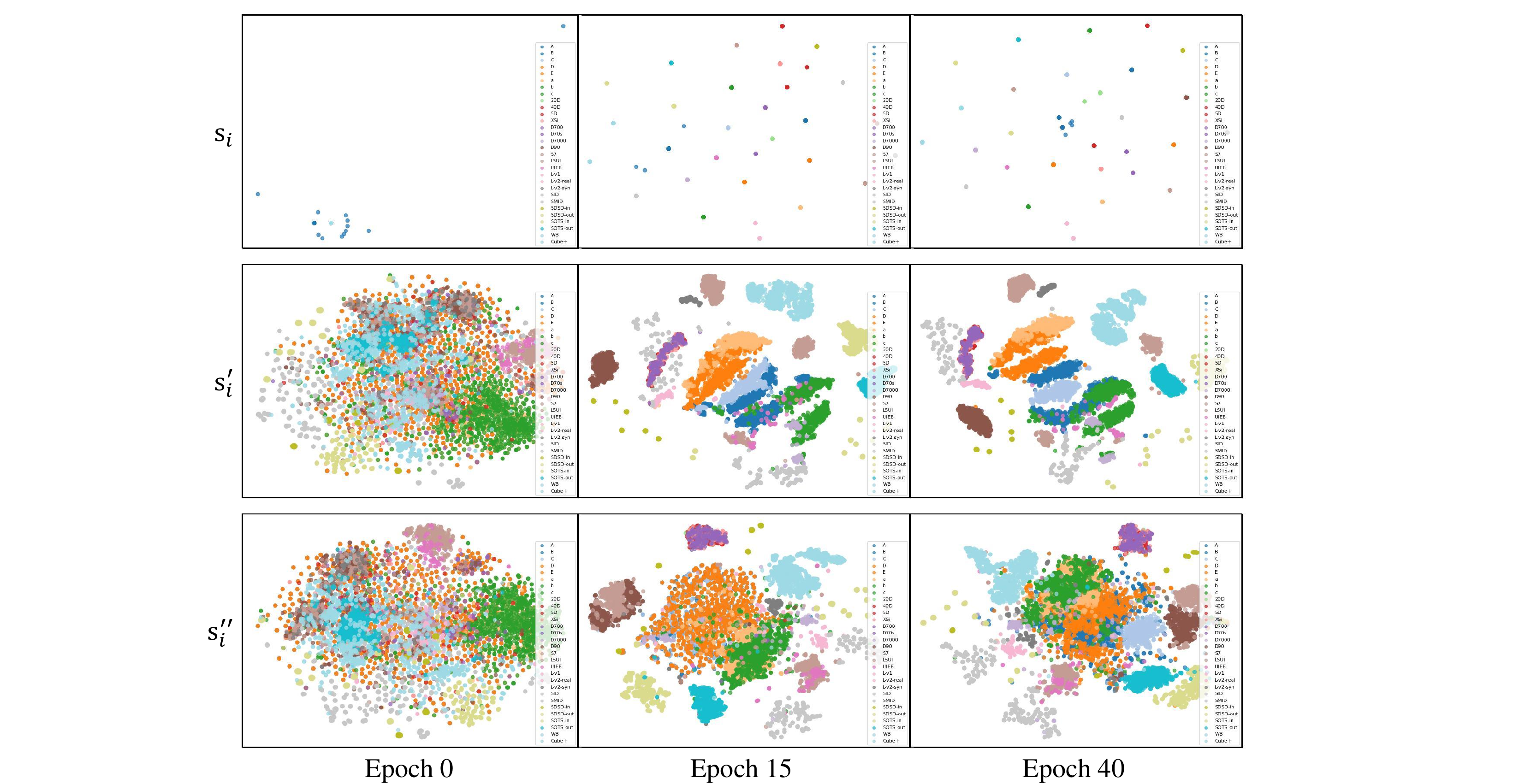}
\caption{Evolution of style tokens $\bfs_i$, $\bfs_i'$, and $\bfs_i''$ during the training.} 
\label{fig:tsne_supp_epoch}
\end{figure*}

\begin{figure}[]
\centering 
\includegraphics[width=1\linewidth]{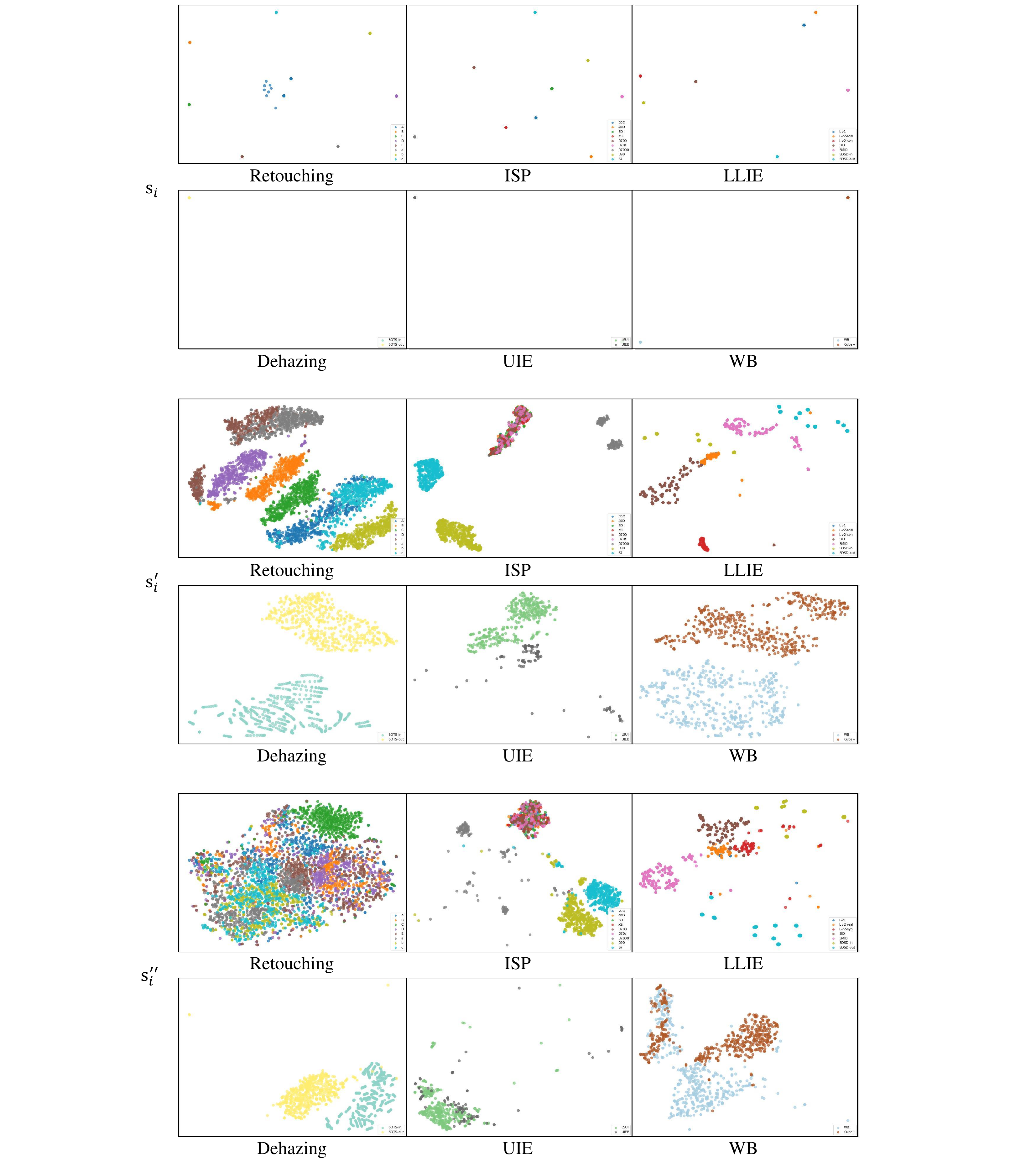}
\caption{t-SNE visualization of style tokens $\bfs_i$, $\bfs_i'$, and $\bfs_i''$ for each task.} 
\label{fig:tsne_supp_task}
\end{figure}

Figure~\ref{fig:tsne_supp_epoch} visualizes how style tokens $\bfs_i$, $\bfs_i'$, and $\bfs_i''$ are aligned through training. Also, Figure \ref{fig:tsne_supp_task} visualizes the style tokens $\bfs_i$, $\bfs_i'$, and $\bfs_i''$ for each of the six tasks separately, after the training.

\clearpage

\subsection{Post-processing for monotonicity}
Because of the eigenTF approximation, a reconstructed TF $\tilde{\bfx}$ in \eqref{eq:backward} may not satisfy the monotonicity constraint. Thus, in the inference phase only, we post-process a TF according to the constraint by setting $x_k=x_{k-1}$ when $x_{k} < x_{k-1}$. Without this post-processing, the performance degrades slightly on some datasets. Table~\ref{table:ablation3} compares the results with and without post-processing.   

\begin{table*}[h]
\centering
\vspace*{0.3cm} 
\renewcommand{\arraystretch}{1.0}
\fontsize{7.5pt}{9pt}\selectfont
\caption{Comparison of the PSNR results with and without post-processing (PP) for the monotonicity of TFs.}
\vspace*{-0.2cm}

\begin{minipage}{1.0\linewidth}  
    \centering
    \fontsize{7.0pt}{8pt}\selectfont
    \resizebox{\textwidth}{!}{
    \begin{tabularx}{\textwidth}{l>{\centering\arraybackslash}X>{\centering\arraybackslash}X>{\centering\arraybackslash}X>{\centering\arraybackslash}X>{\centering\arraybackslash}X>{\centering\arraybackslash}X>{\centering\arraybackslash}X>{\centering\arraybackslash}X}
    \toprule
    {\textbf{Retouching}} & A & B & C & D & E & a & b & c \\
    \midrule
    w/o PP & 22.27 & 25.37 & 23.91 & 22.51 & 23.40 & 21.80 & 21.57 & 22.66 \\
    w/ PP & 22.27 & 25.37 & 23.91 & 22.51 & 23.40 & 21.80 & 21.57 & 22.66 \\
    \bottomrule
    \end{tabularx}}
\end{minipage}

\vspace*{0.1cm} 

\begin{minipage}{1.0\linewidth}  
    \centering
    \fontsize{7.0pt}{8pt}\selectfont
    \resizebox{\textwidth}{!}{
    \begin{tabularx}{\textwidth}{l>{\centering\arraybackslash}X>{\centering\arraybackslash}X>{\centering\arraybackslash}X>{\centering\arraybackslash}X>{\centering\arraybackslash}X>{\centering\arraybackslash}X>{\centering\arraybackslash}X>{\centering\arraybackslash}X>{\centering\arraybackslash}X}

    \toprule
    {\textbf{ISP}} & 20D & 40D & 5D & XSi & D700 & D70s & D7000 & D90 & S7\\
    \midrule
    w/o PP & 29.21 & 28.53 & 29.31 & 29.38 & 29.50 & 28.89 & 23.86 & 22.28 & 24.37 \\
    w/ PP & 29.21 & 28.53 & 29.31 & 29.38 & 29.69 & 28.89 & 23.86 & 22.28 & 24.37 \\
    \bottomrule
    \end{tabularx}}
\end{minipage}

\vspace*{0.1cm} 

\begin{minipage}{1.0\linewidth}  
    \centering
    \fontsize{7.0pt}{8pt}\selectfont
    \resizebox{\textwidth}{!}{
    \begin{tabularx}{\textwidth}{l>{\centering\arraybackslash}X>{\centering\arraybackslash}X>{\centering\arraybackslash}X>{\centering\arraybackslash}X>{\centering\arraybackslash}X>{\centering\arraybackslash}X>{\centering\arraybackslash}X}
    \toprule
    {\textbf{LLIE}} & L-v1 & L-v2-real & L-v2-syn & SID & SMID & SDSD-in & SDSD-out \\
    \midrule
    w/o PP & 24.86 & 27.33 & 22.15 & 18.23 & 25.56 & 27.06 & 28.19 \\
    w/ PP & 24.89 & 27.91 & 22.15 & 18.23 & 25.56 & 27.06 & 28.19 \\
    \bottomrule
    \end{tabularx}}
\end{minipage}

\vspace*{0.1cm} 

\hspace{0.01em}
\begin{minipage}{0.320\linewidth}
    \centering
    \fontsize{7.0pt}{8pt}\selectfont
    \begin{tabularx}{\linewidth}{l>{\centering\arraybackslash}X>{\centering\arraybackslash}X}
        \toprule
        {\textbf{Dehazing}} & SOTS-in & SOTS-out  \\
        \midrule
        w/o PP & 20.66 & 28.26 \\  
        w/ PP & 20.66 & 28.51 \\
        \bottomrule
    \end{tabularx}
\end{minipage}
\hspace{0.6em}
\begin{minipage}{0.320\linewidth}
    \centering
    \fontsize{7.0pt}{8pt}\selectfont
    \begin{tabularx}{\linewidth}{l>{\centering\arraybackslash}X>{\centering\arraybackslash}X}
        \toprule
         {\textbf{UIE}}  & LSUI & UIEB \\
        \midrule
        w/o PP & 24.90 & 22.01(19.87) \\
        w/ PP & 24.90 & 22.15(19.87) \\
        \bottomrule
    \end{tabularx}
\end{minipage}
\hspace{0.6em}
\begin{minipage}{0.320\linewidth}
    \centering
    \fontsize{7.0pt}{8pt}\selectfont
    \begin{tabularx}{\linewidth}{l>{\centering\arraybackslash}X>{\centering\arraybackslash}X}
        \toprule
        {\textbf{WB}} & WB & Cube+  \\
        \midrule
        w/o PP & 27.08 & 29.41(28.92) \\
        w/ PP & 27.10 & 29.54(28.92) \\
        \bottomrule
    \end{tabularx}
\end{minipage}
\label{table:ablation3}
\end{table*}

\clearpage

\section{Oneta enhancement results}

\begin{figure*}[h]
\centering 
\includegraphics[width=1\linewidth]{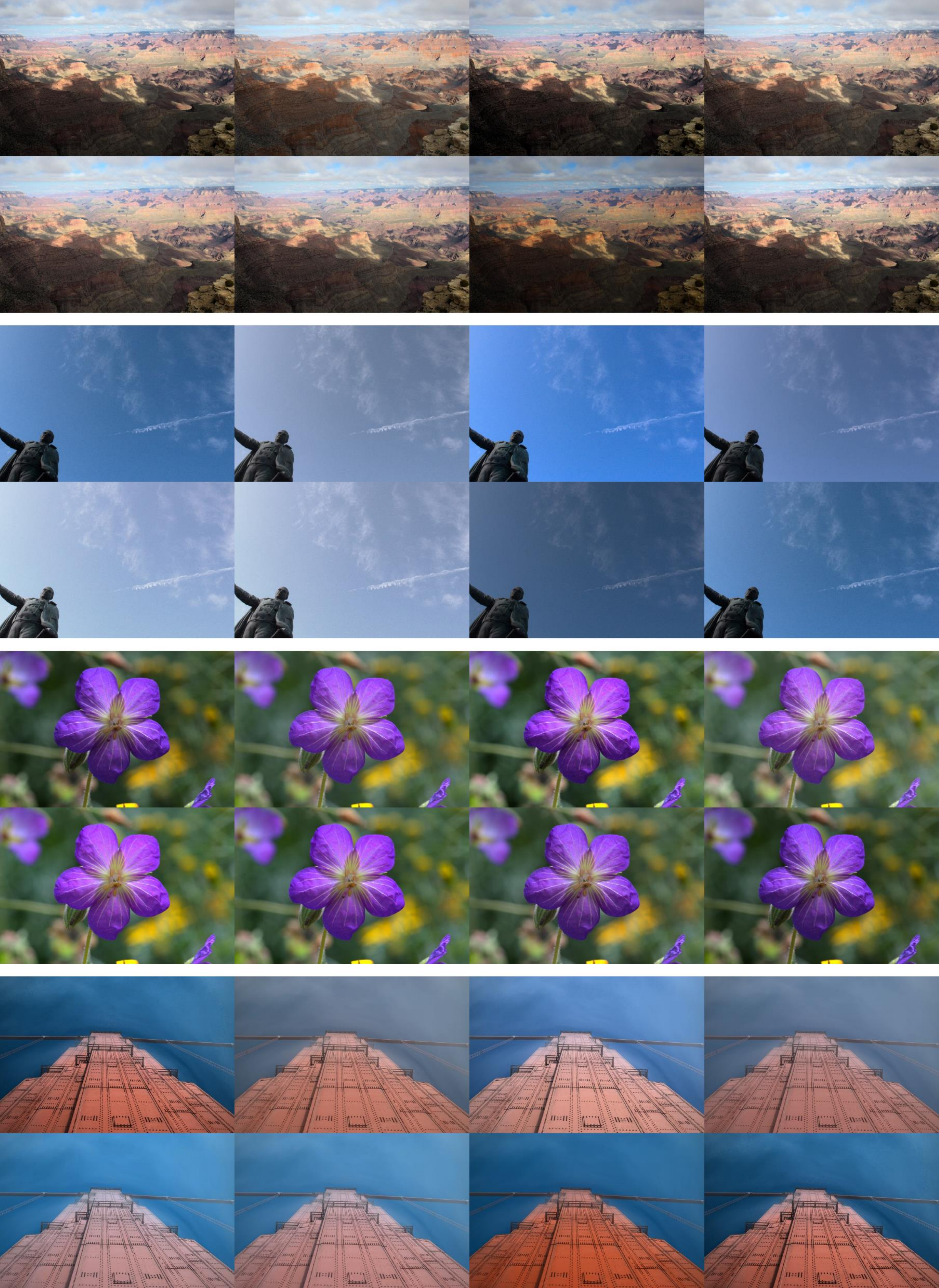}
\caption{Retouching examples on the FiveK dataset. For each image, the retouching results of expert styles A, B, C, D, E, a, b, and c are shown in the raster scan order.}
\vspace{-0.4cm}
\label{fig:supp_qual_mit}
\end{figure*}

\begin{figure*}[]
\centering 
\includegraphics[width=1\linewidth]{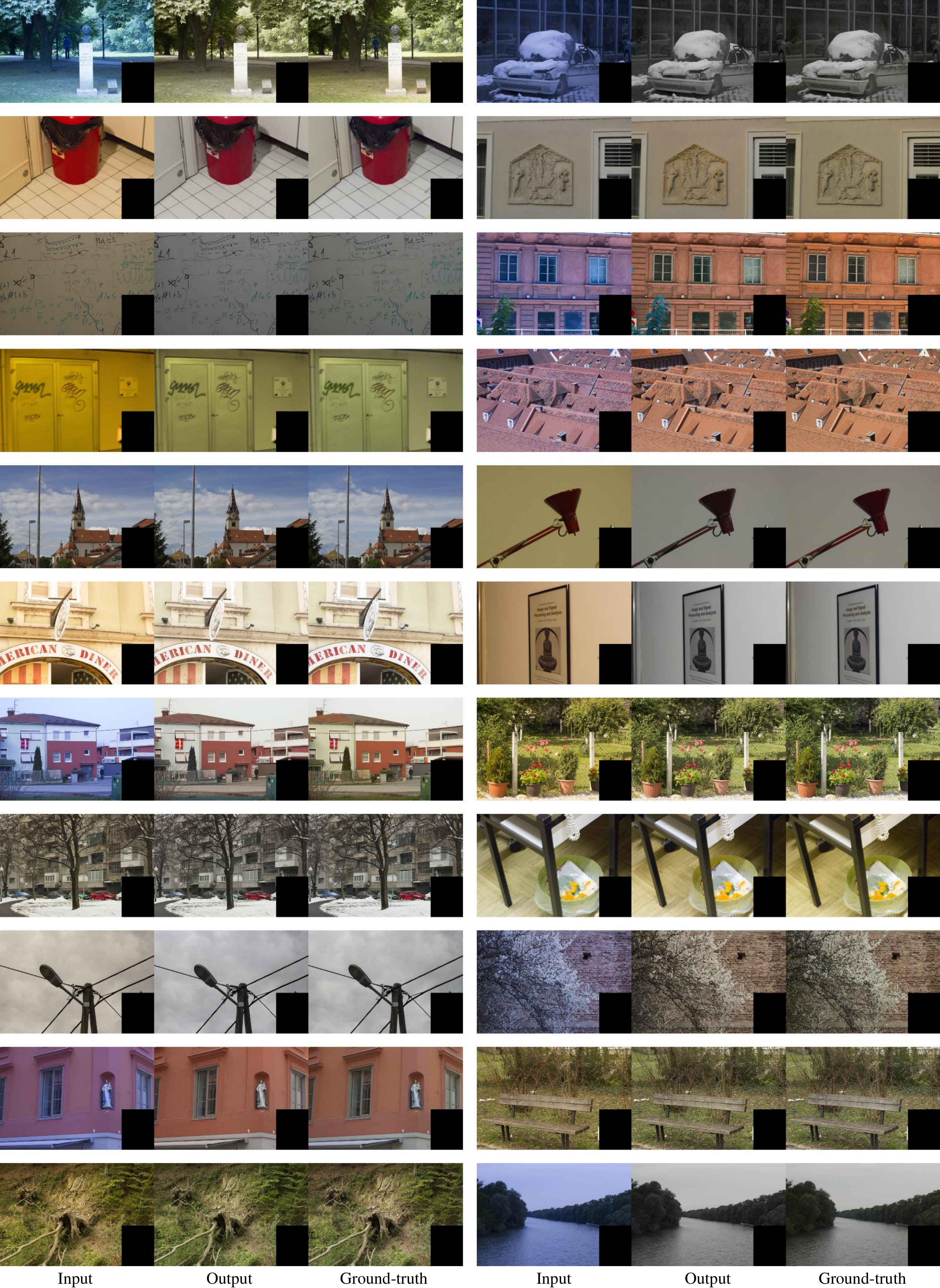}
\caption{WB results on the Cube+ dataset.} 
\label{fig:supp_qual_29}
\end{figure*}

Figures \ref{fig:supp_qual_mit} $ \sim $ \ref{fig:supp_qual_29} show Oneta enhancement examples on the 30 datasets. 

\subsection{Retouching}
\vspace{-0.4cm}

\begin{figure*}[]
\centering 
\includegraphics[width=1\linewidth]{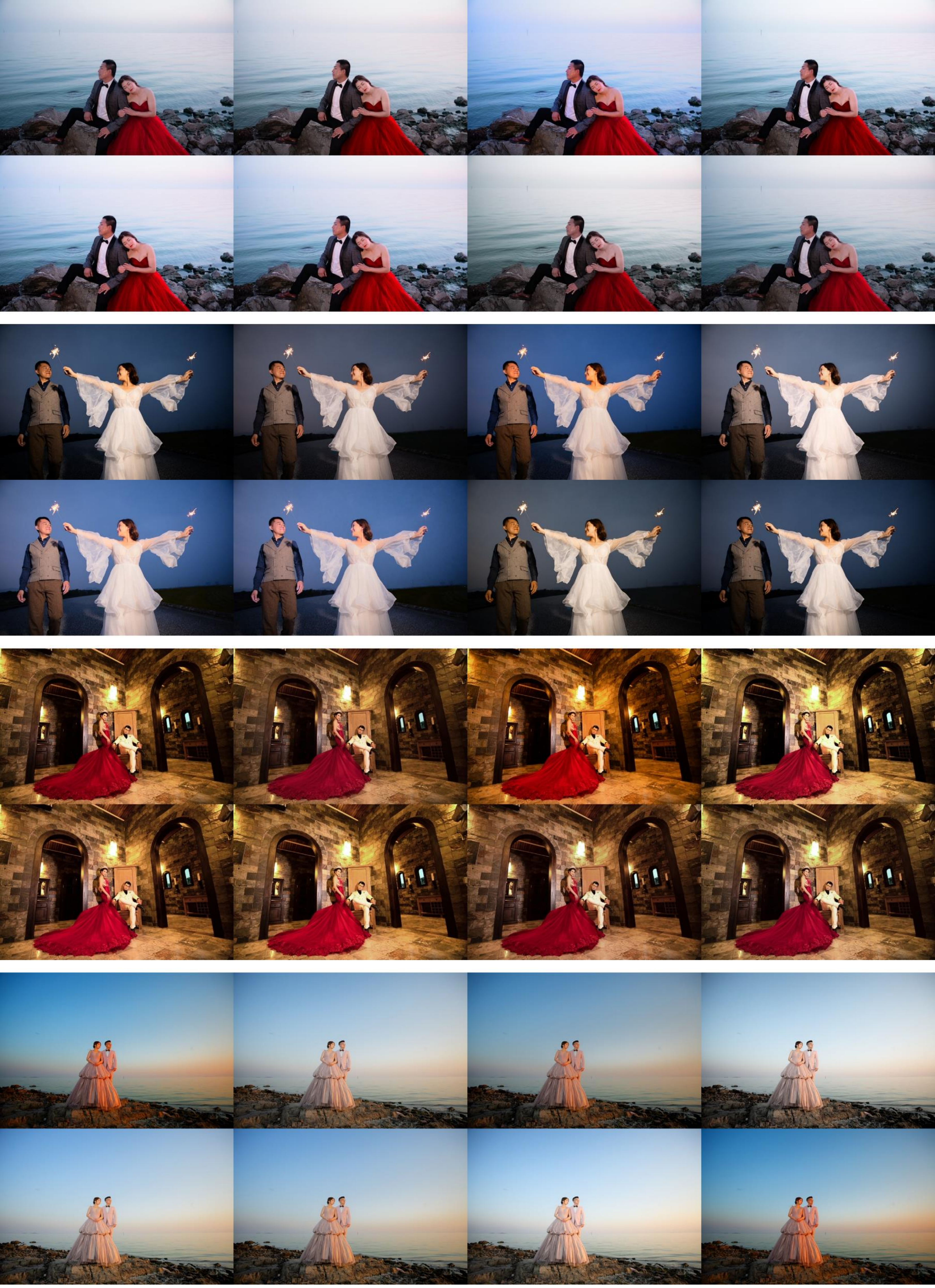}
\caption{Retouching examples on the PPR10K dataset.  For each image, the retouching results of expert styles A, B, C, D, E, a, b, and c are shown in the raster scan order.}
\end{figure*}

\clearpage

\subsection{Image Signal Processing}

\begin{figure*}[h]
\centering 
\includegraphics[width=1\linewidth]{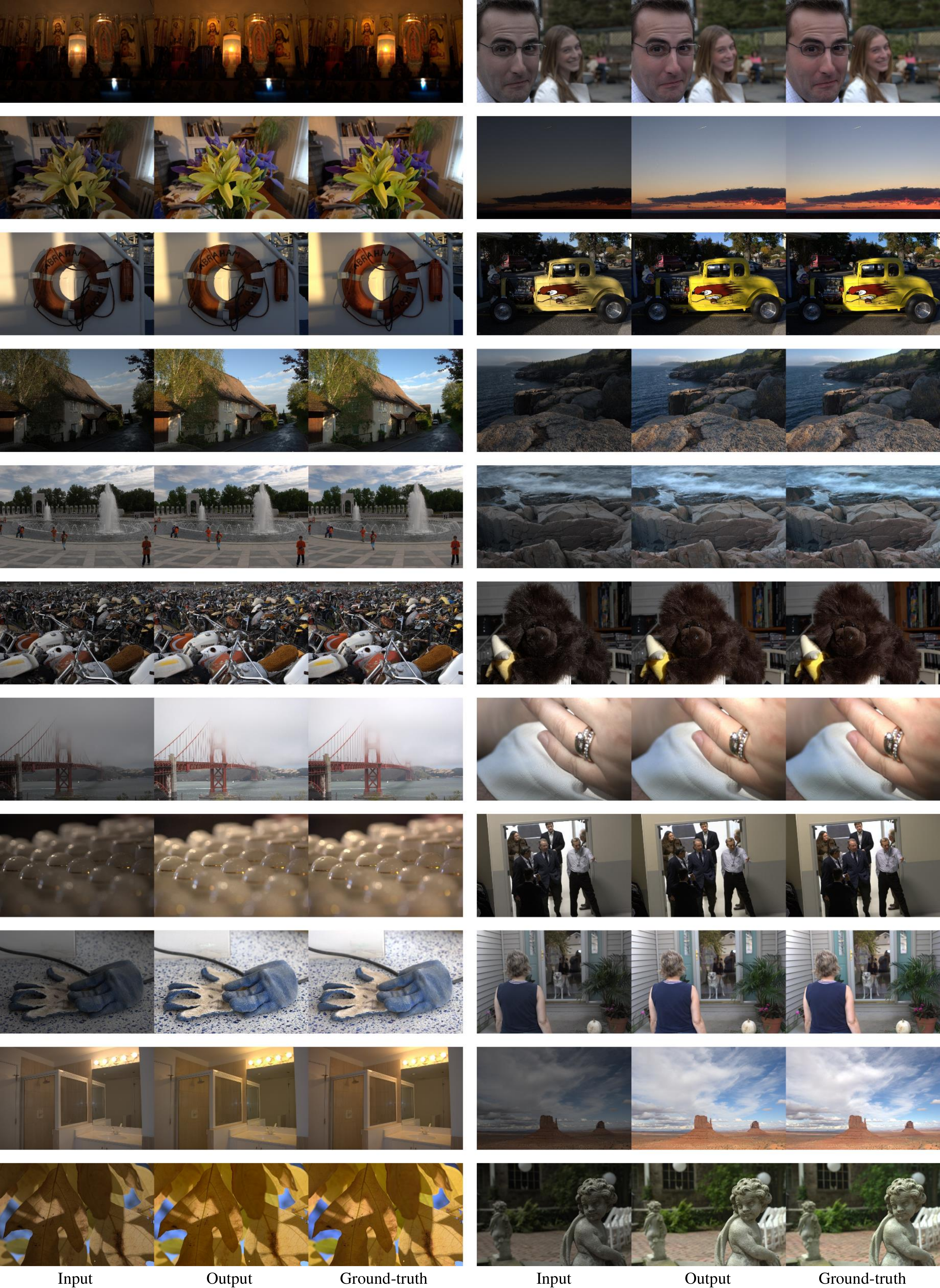}
\caption{ISP results on the FiveK-Canon EOS 20D dataset.}
\end{figure*}

\begin{figure*}[]
\centering 
\includegraphics[width=1\linewidth]{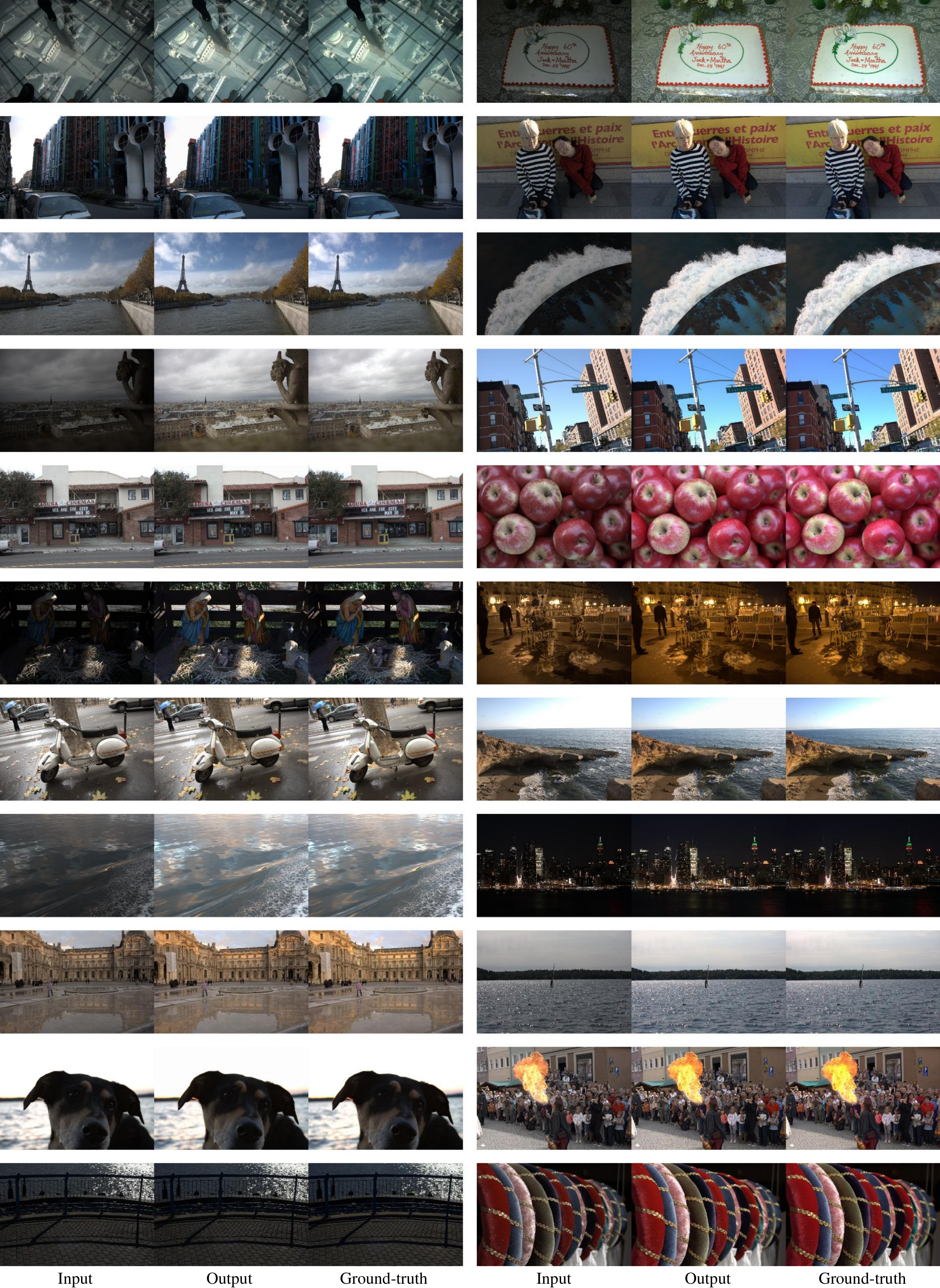}
\caption{ISP results on the FiveK-Canon EOS 40D dataset.} 
\end{figure*}

\begin{figure*}[]
\centering 
\includegraphics[width=1\linewidth]{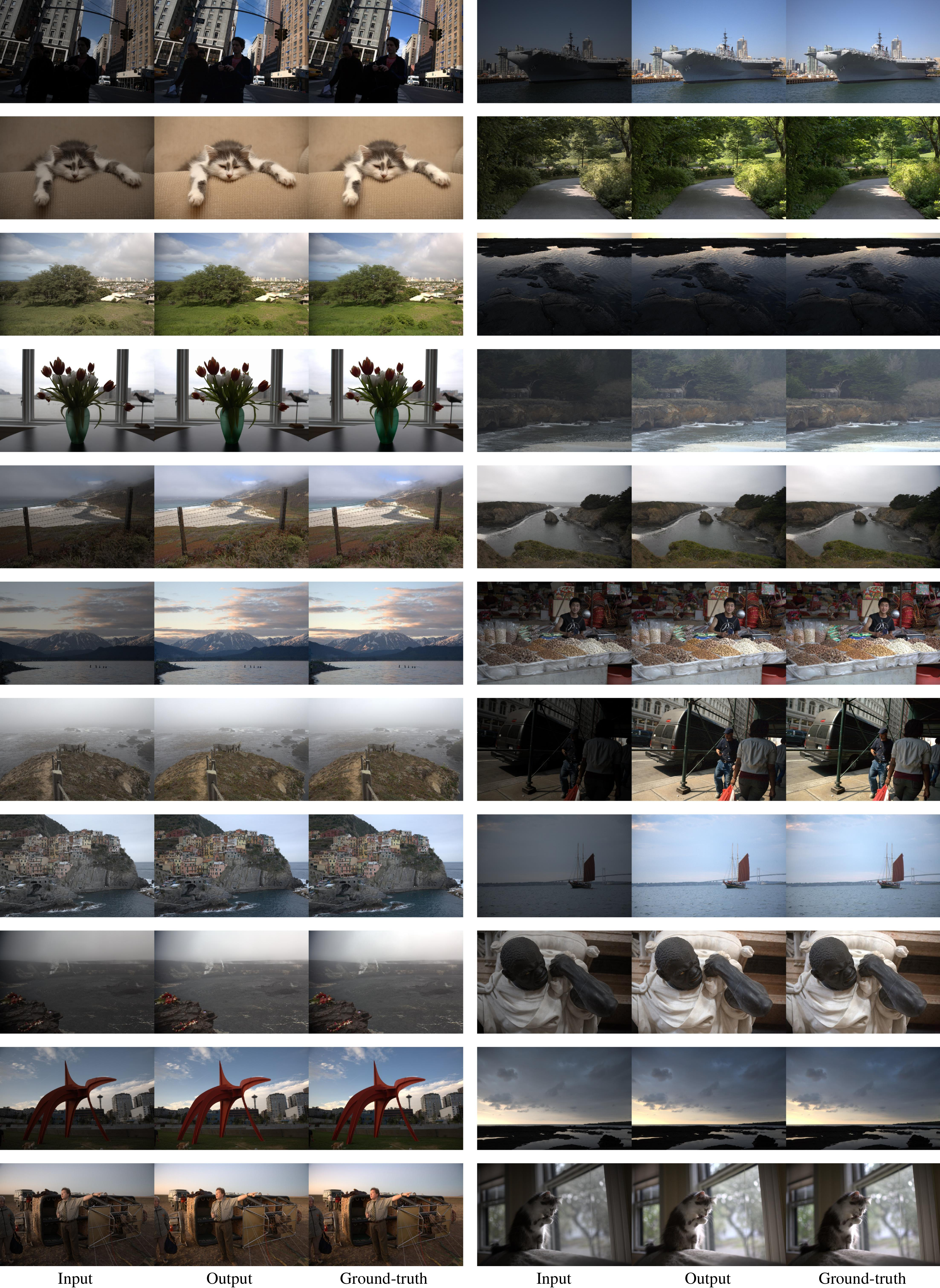}
\caption{ISP results on the FiveK-Canon EOS 5D dataset.} 
\end{figure*}

\begin{figure*}[]
\centering 
\includegraphics[width=1\linewidth]{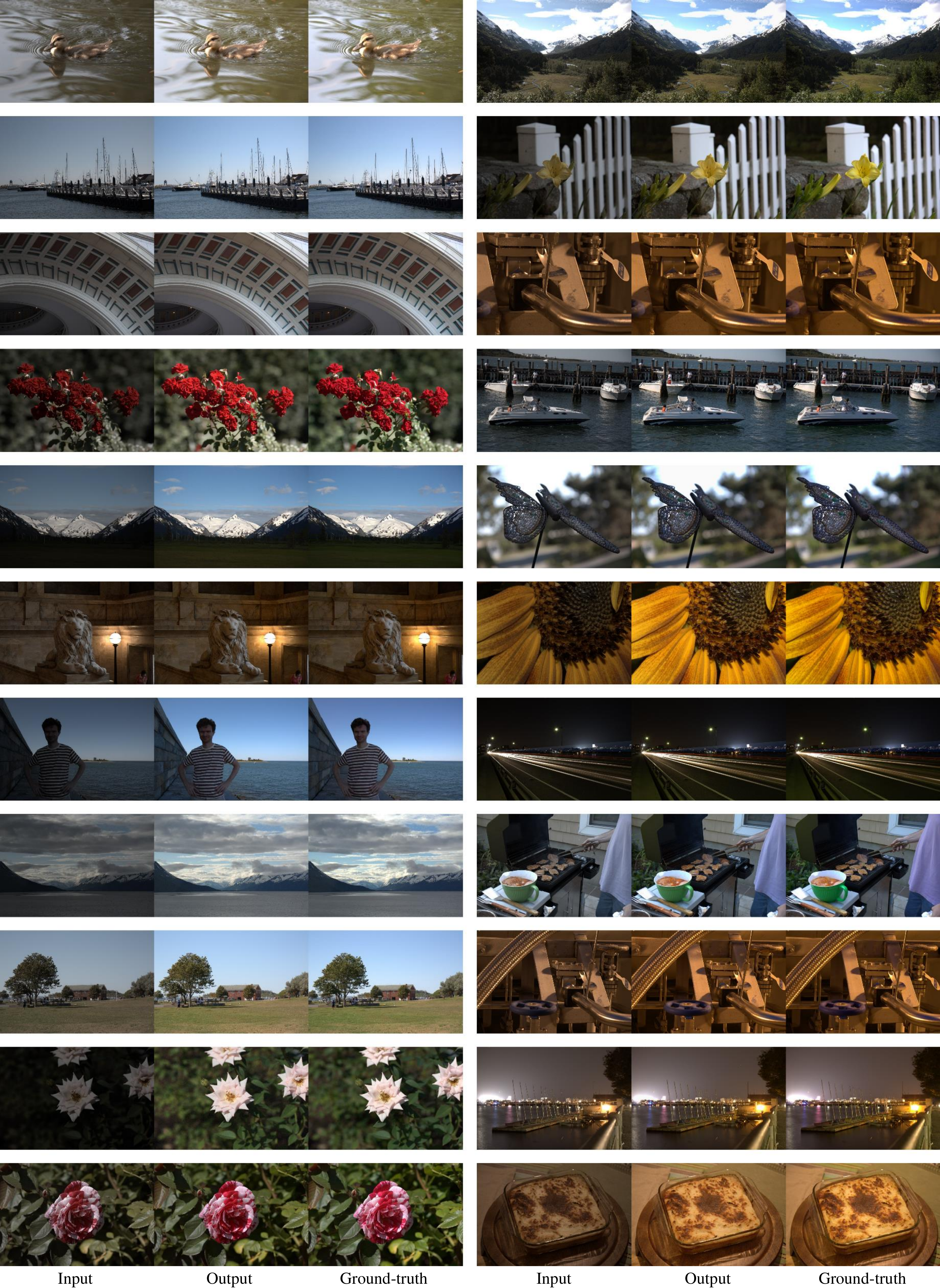}
\caption{ISP results on the FiveK-Canon EOS XSi dataset.} 
\end{figure*}

\begin{figure*}[]
\centering 
\includegraphics[width=1\linewidth]{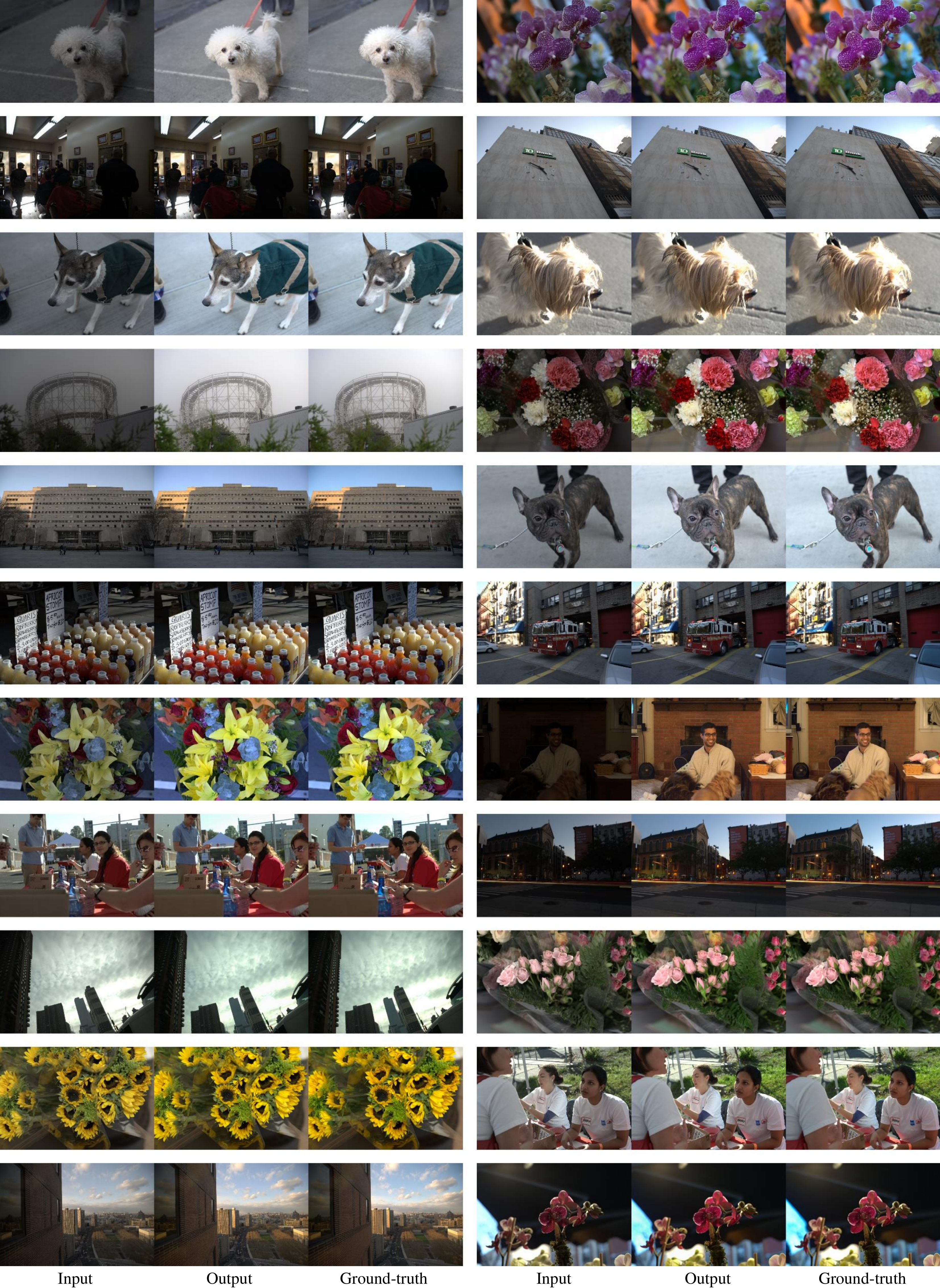}
\caption{ISP results on the FiveK-Nikon D700 dataset.} 
\end{figure*}

\begin{figure*}[]
\centering 
\includegraphics[width=1\linewidth]{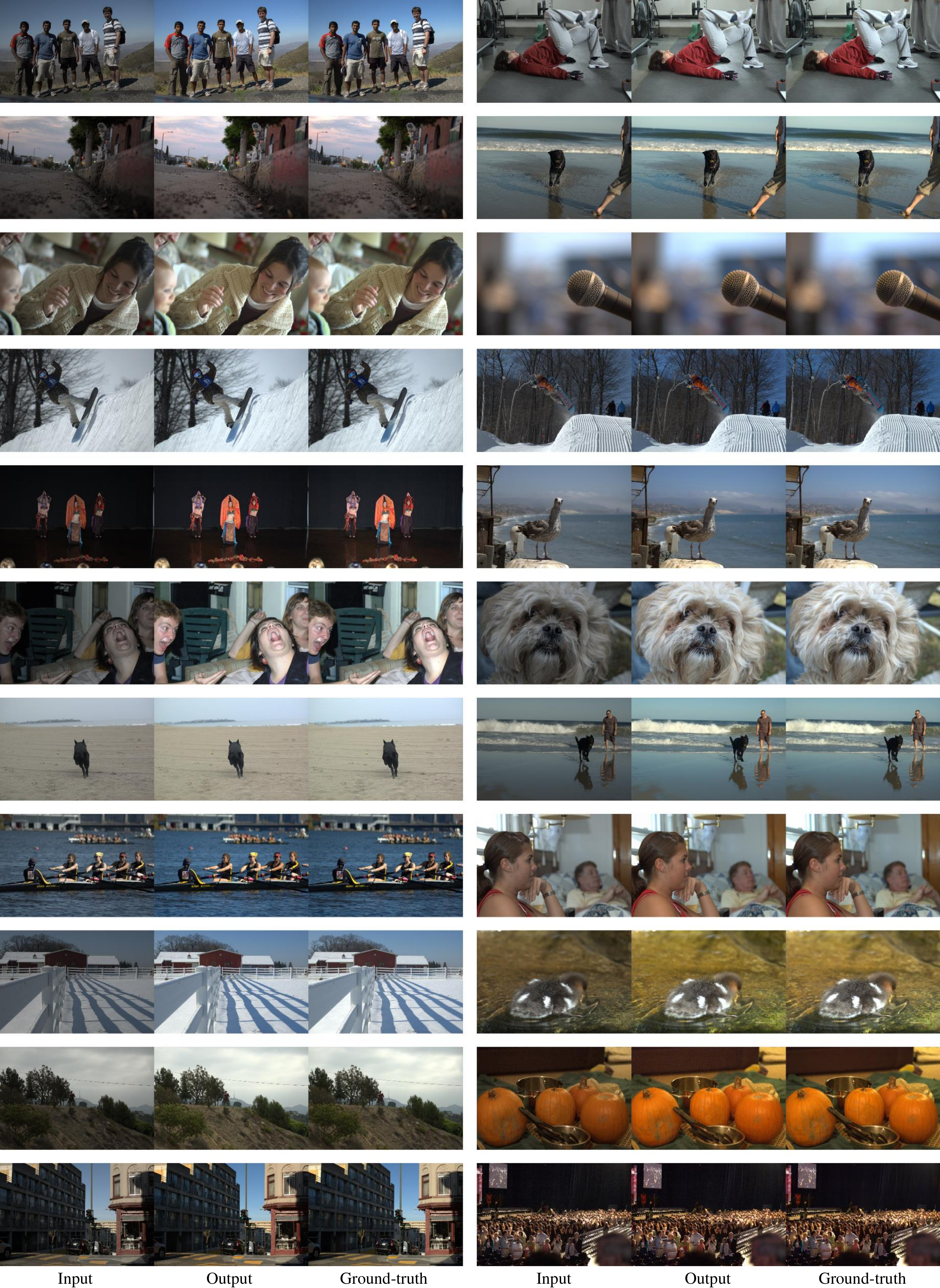}
\caption{ISP results on the FiveK-Nikon D70s dataset.} 
\end{figure*}

\begin{figure*}[]
\centering 
\includegraphics[width=1\linewidth]{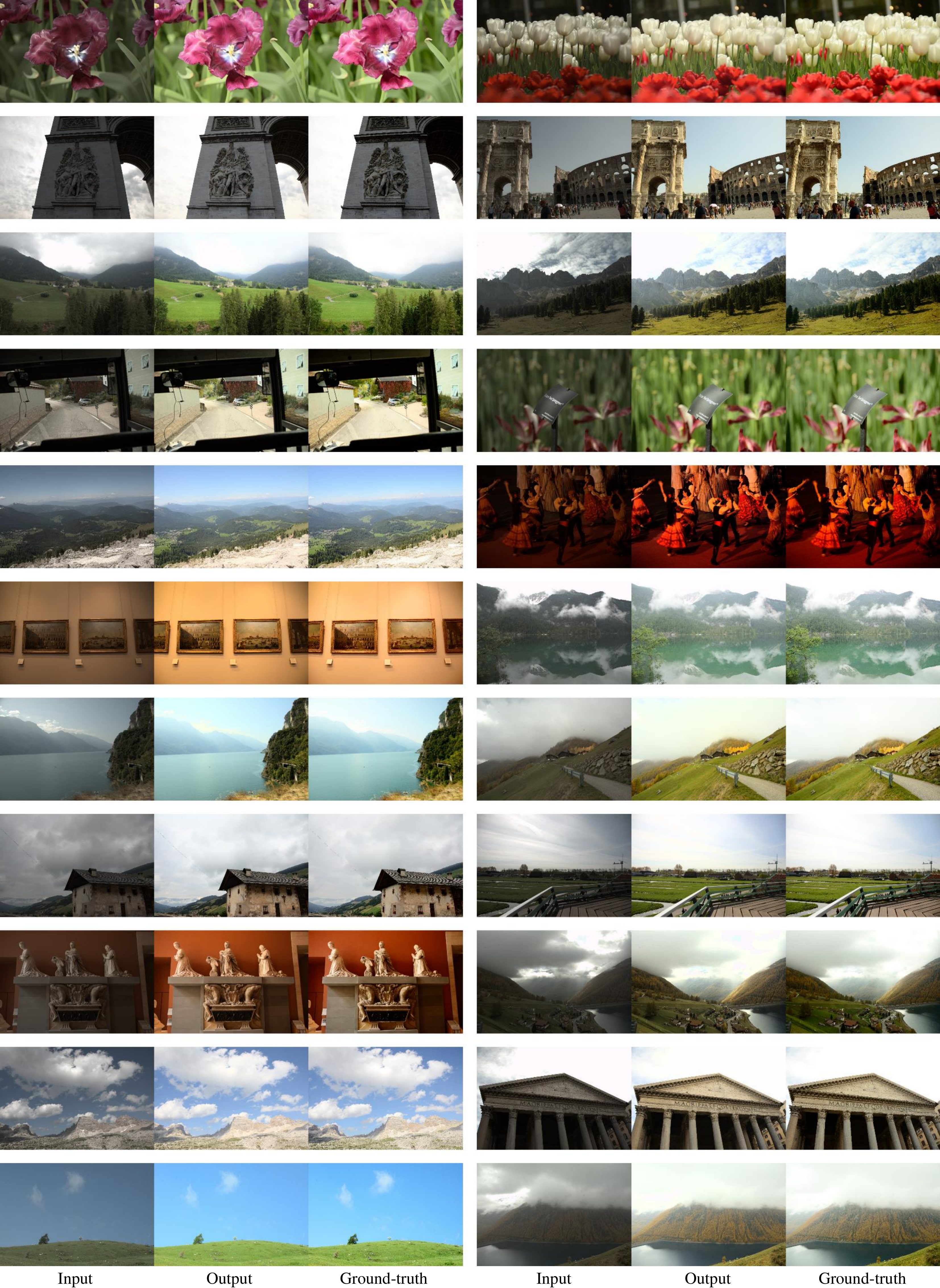}
\caption{ISP results on the RAISE-Nikon D7000 dataset.} 
\end{figure*}

\begin{figure*}[]
\centering 
\includegraphics[width=1\linewidth]{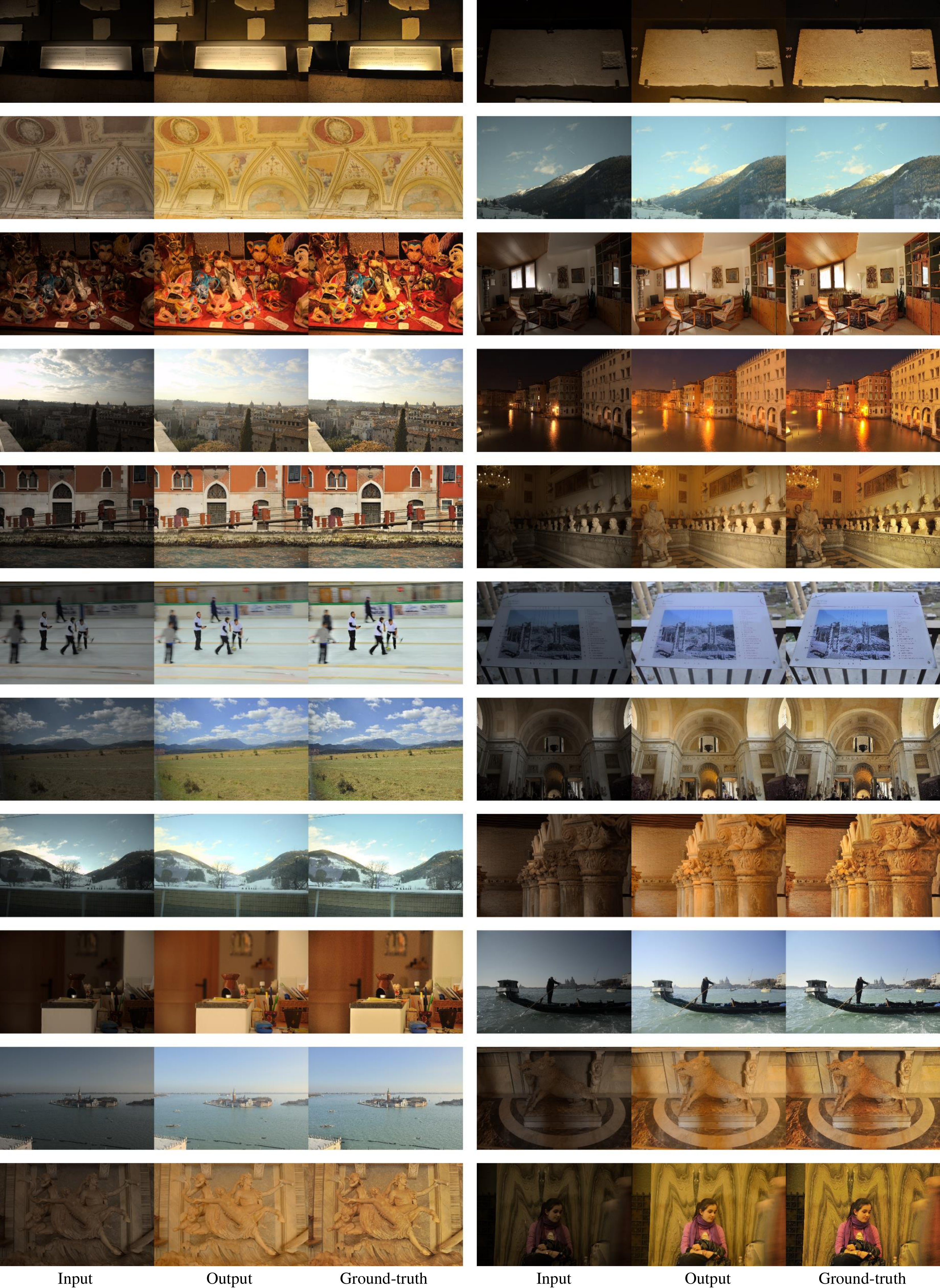}
\caption{ISP results on the Samsung Galaxy D90 dataset.} 
\end{figure*}

\begin{figure*}[]
\centering 
\includegraphics[width=1\linewidth]{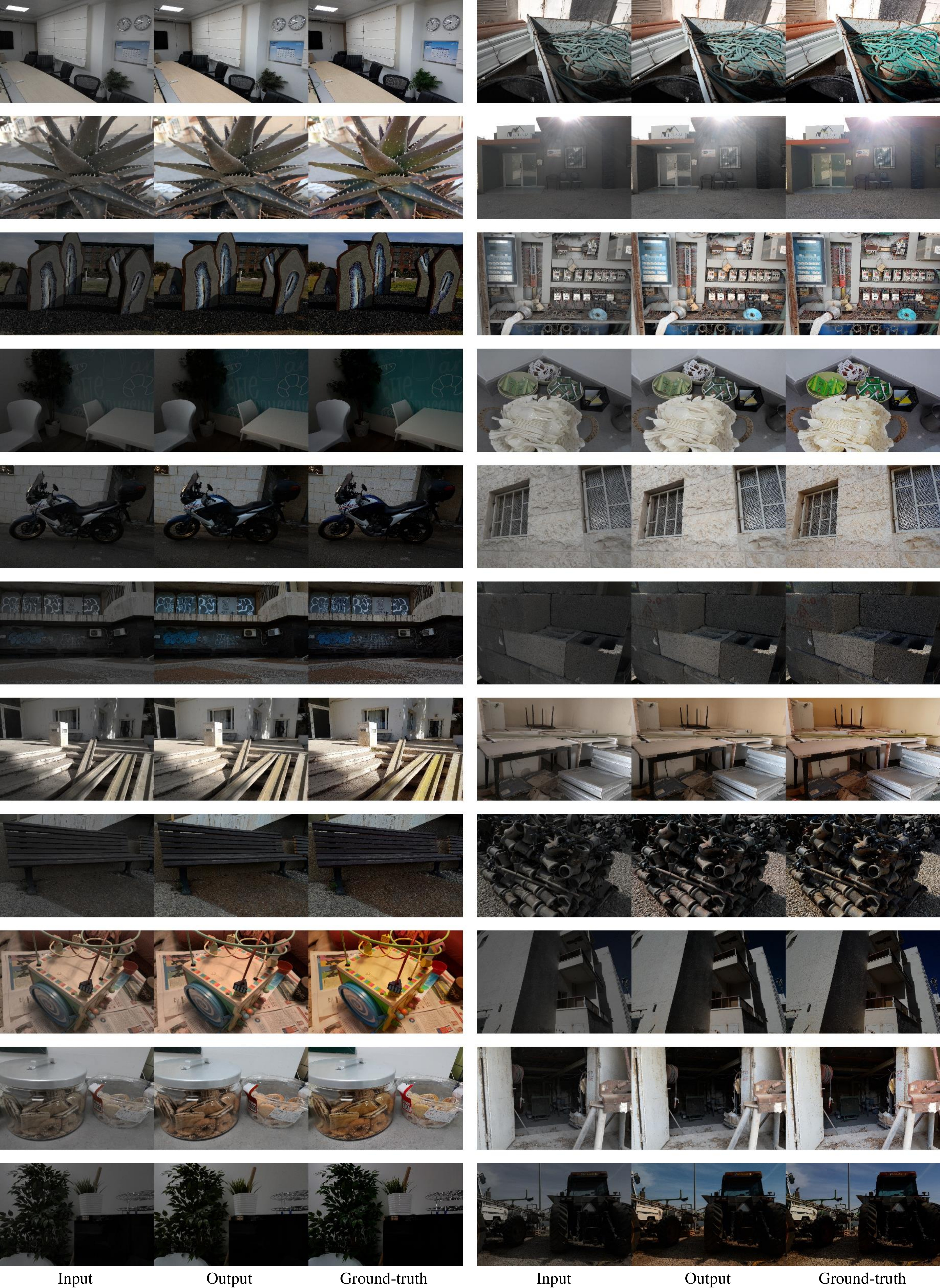}
\caption{ISP results on the RAISE-Nikon S7 dataset.} 
\end{figure*}

\clearpage

\subsection{Low-light Image Enhancement}
\begin{figure*}[h]
\centering 
\includegraphics[width=1\linewidth]{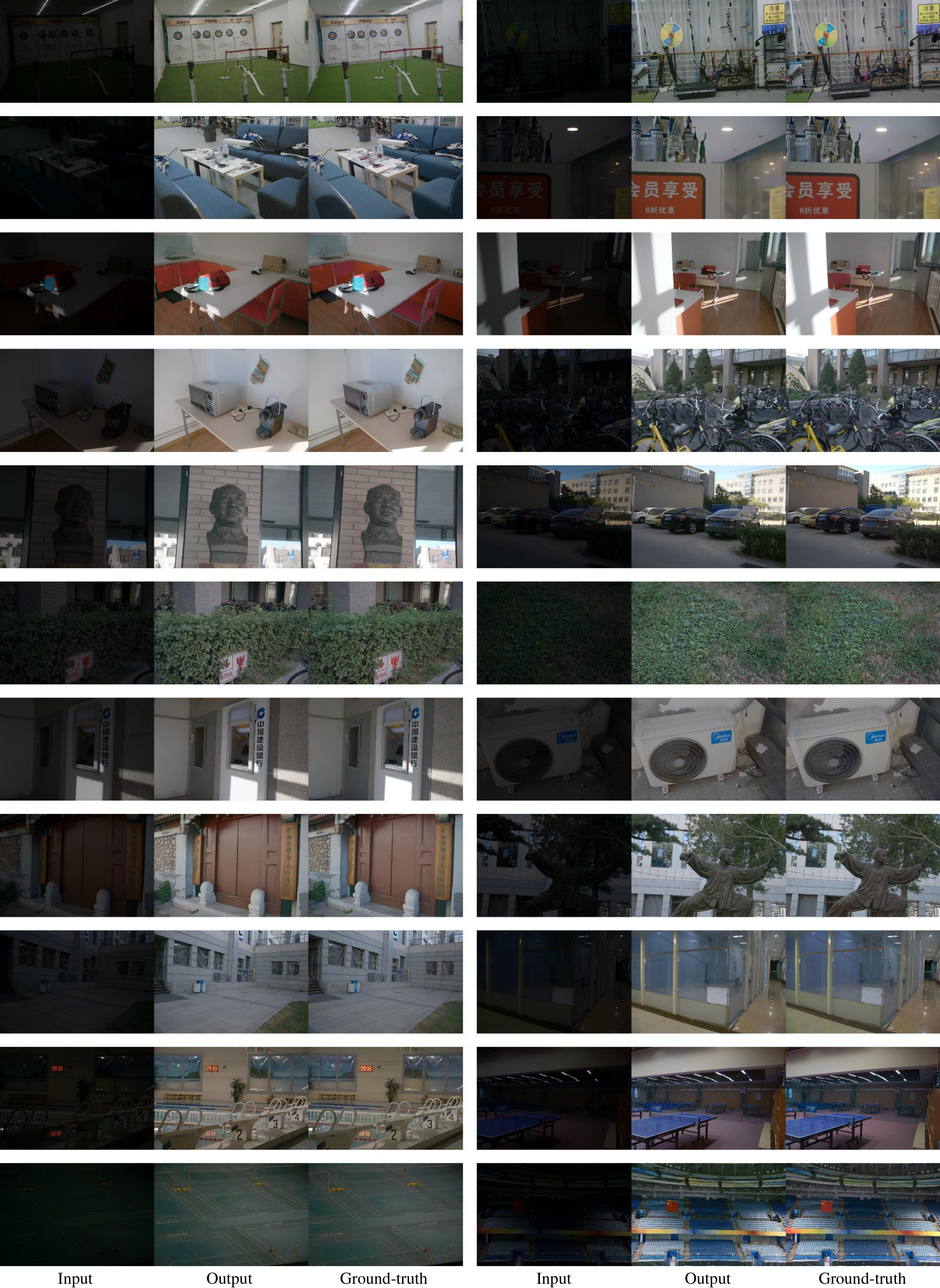}
\caption{LLIE results on the LOL-v2-real dataset.} 
\end{figure*}

\begin{figure*}[]
\centering 
\includegraphics[width=1\linewidth]{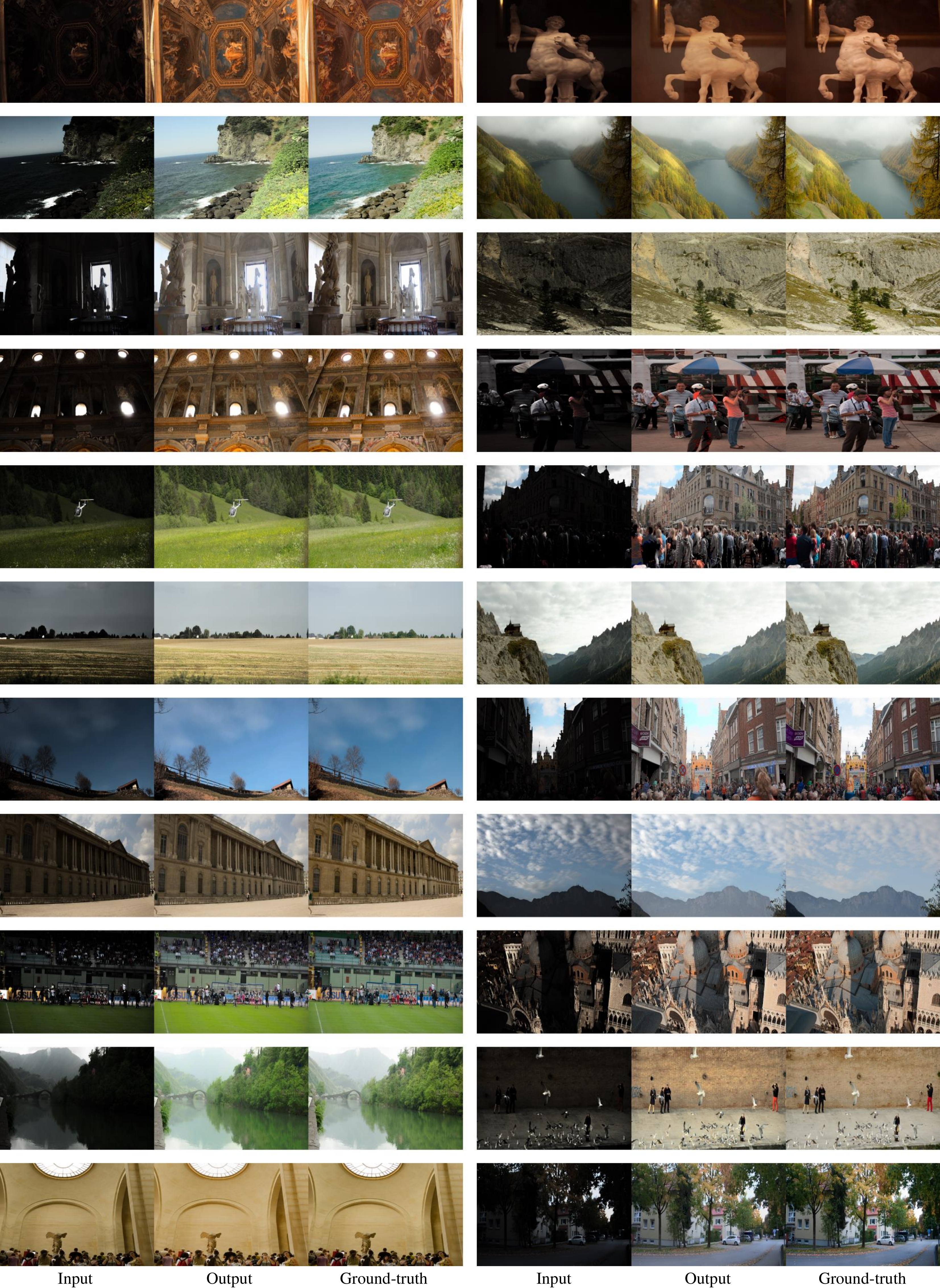}
\caption{LLIE results on the LOL-v2-syn dataset.} 
\end{figure*}

\clearpage
\begin{figure*}[t]
\centering 
\includegraphics[width=1\linewidth]{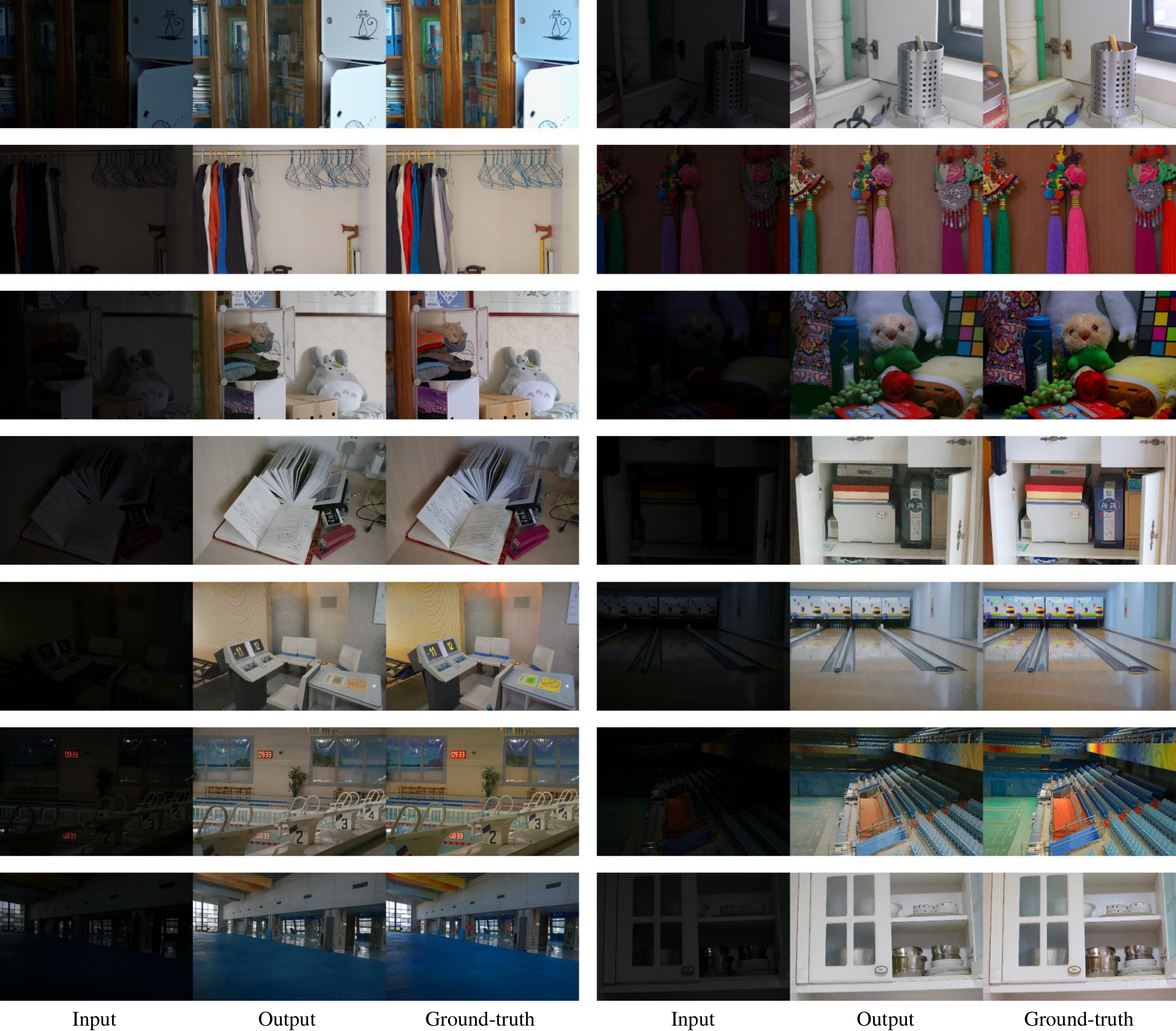}
\captionsetup{aboveskip=5pt}
\caption{LLIE results on the LOL-v1 dataset.} 
\end{figure*}
\vspace{-3cm}
\begin{figure*}[h]
\vspace{-0.4cm}
\centering 
\includegraphics[width=1\linewidth]{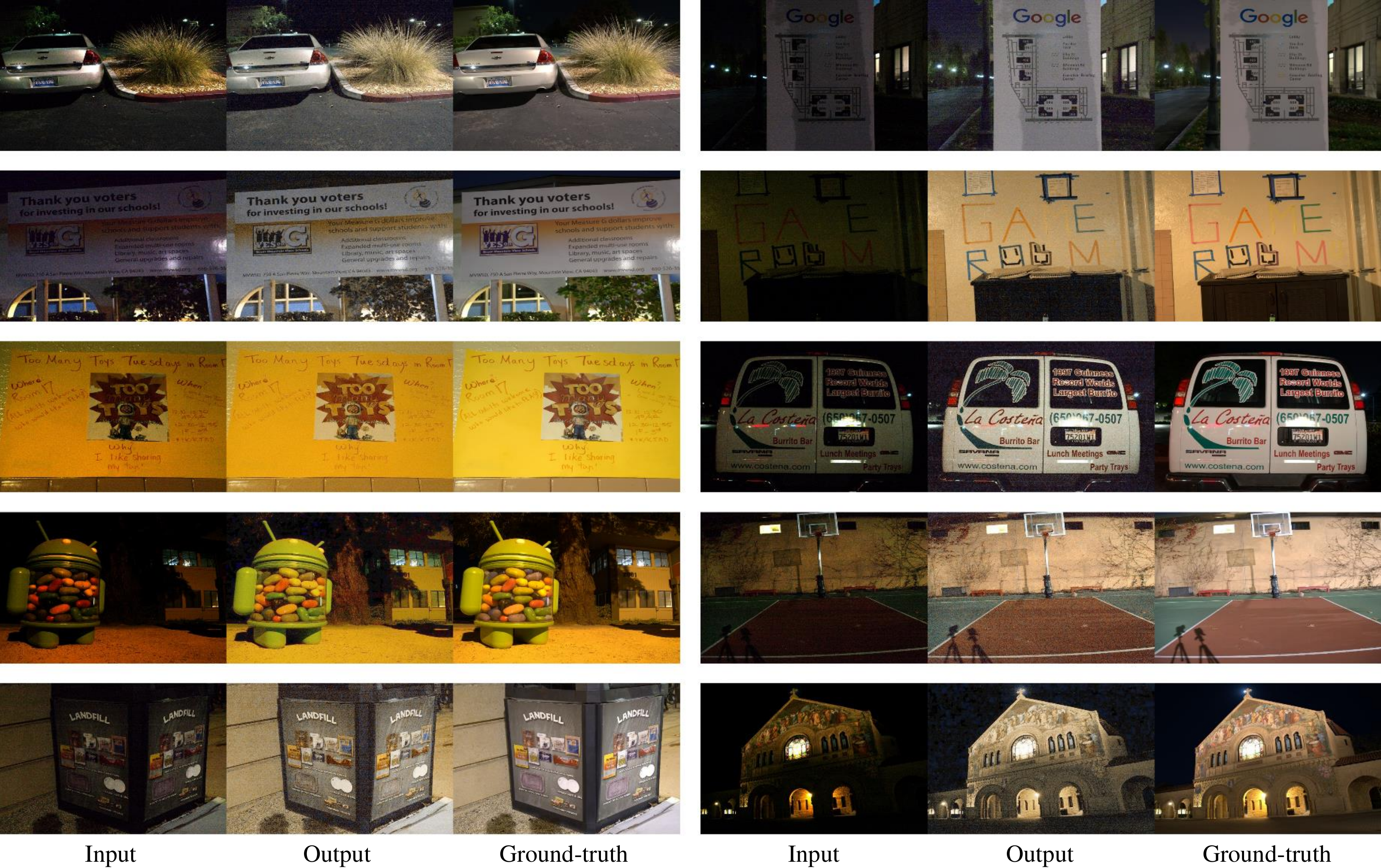}
\captionsetup{aboveskip=5pt}
\caption{LLIE results on the SID dataset.} 
\vspace{-3cm}
\end{figure*}

\begin{figure*}[]
\centering 
\includegraphics[width=1\linewidth]{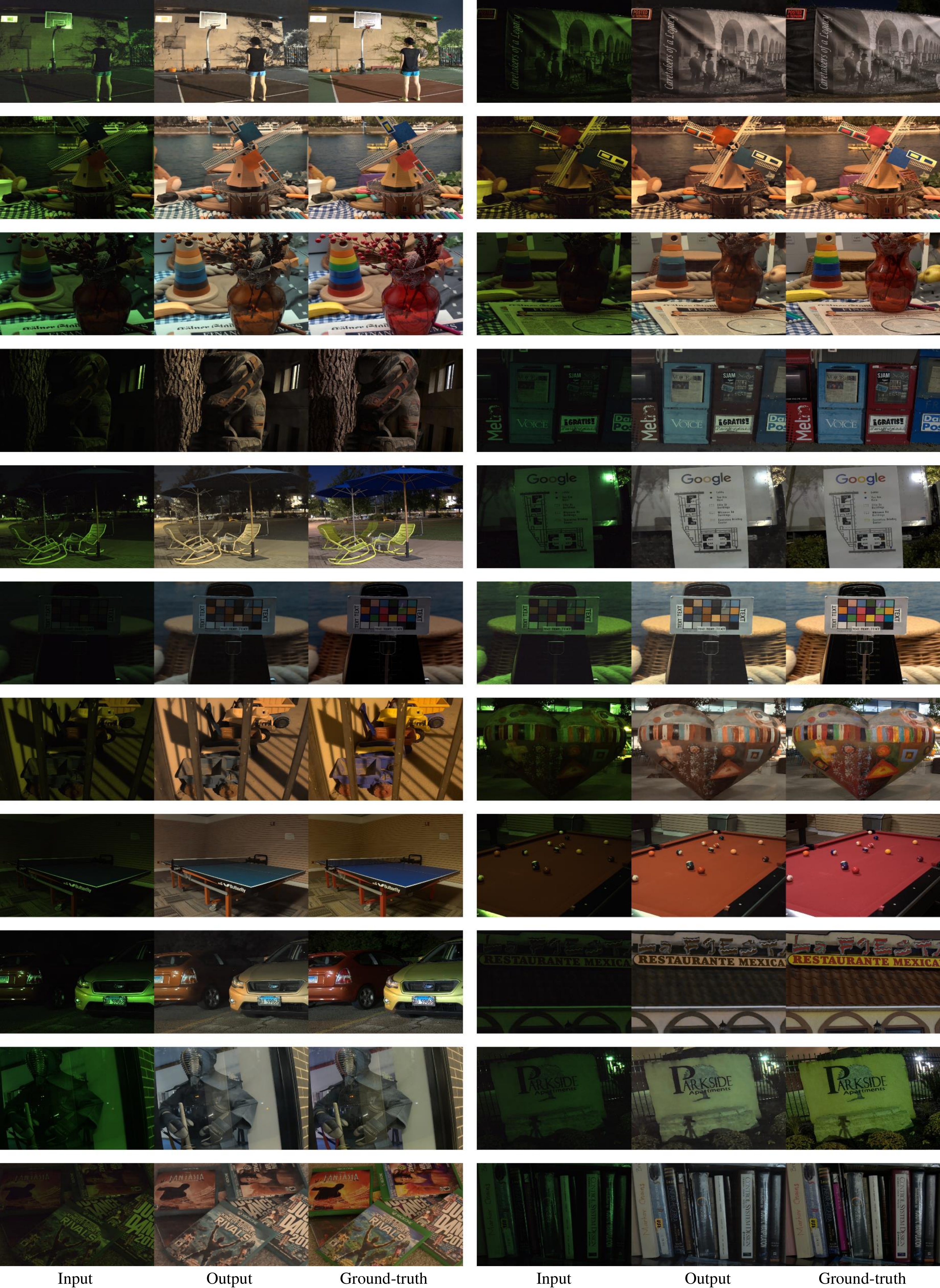}
\caption{LLIE results on the SMID dataset.} 
\end{figure*}

\begin{figure*}[h]
\centering 
\includegraphics[width=1\linewidth]{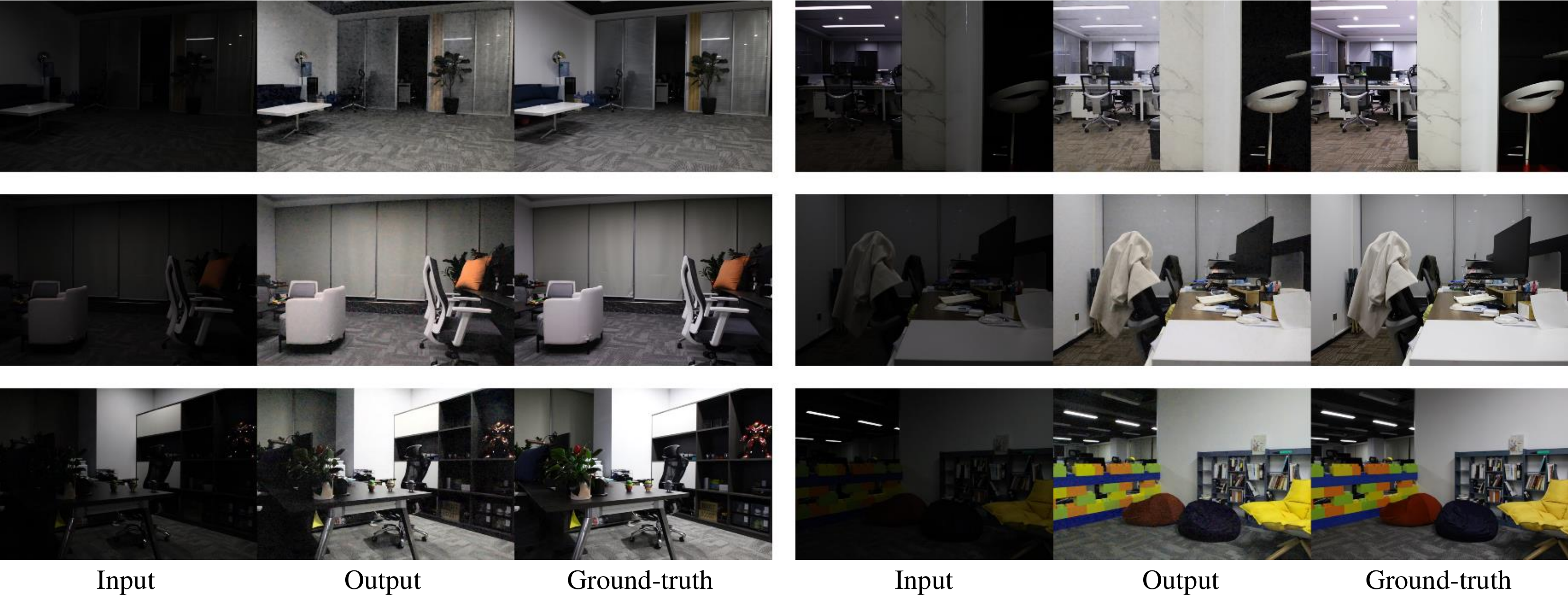}
\caption{LLIE results on the SDSD-indoor dataset.} 
\end{figure*}

\begin{figure*}[h]
\centering 
\includegraphics[width=1\linewidth]{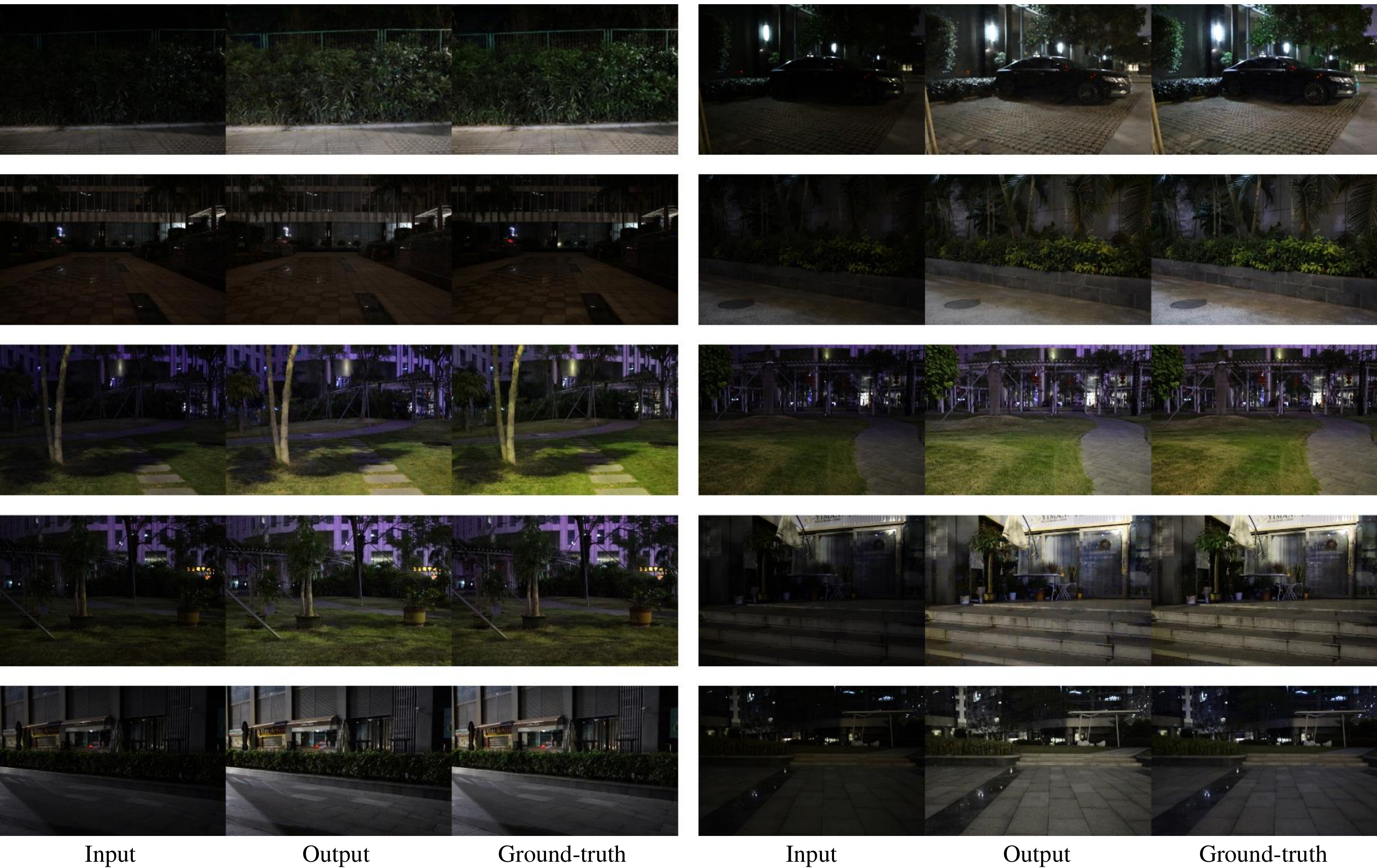}
\caption{LLIE results on the SDSD-outdoor dataset.} 
\end{figure*}

\clearpage

\subsection{Dehazing}
\begin{figure*}[h]
\centering 
\includegraphics[width=1\linewidth]{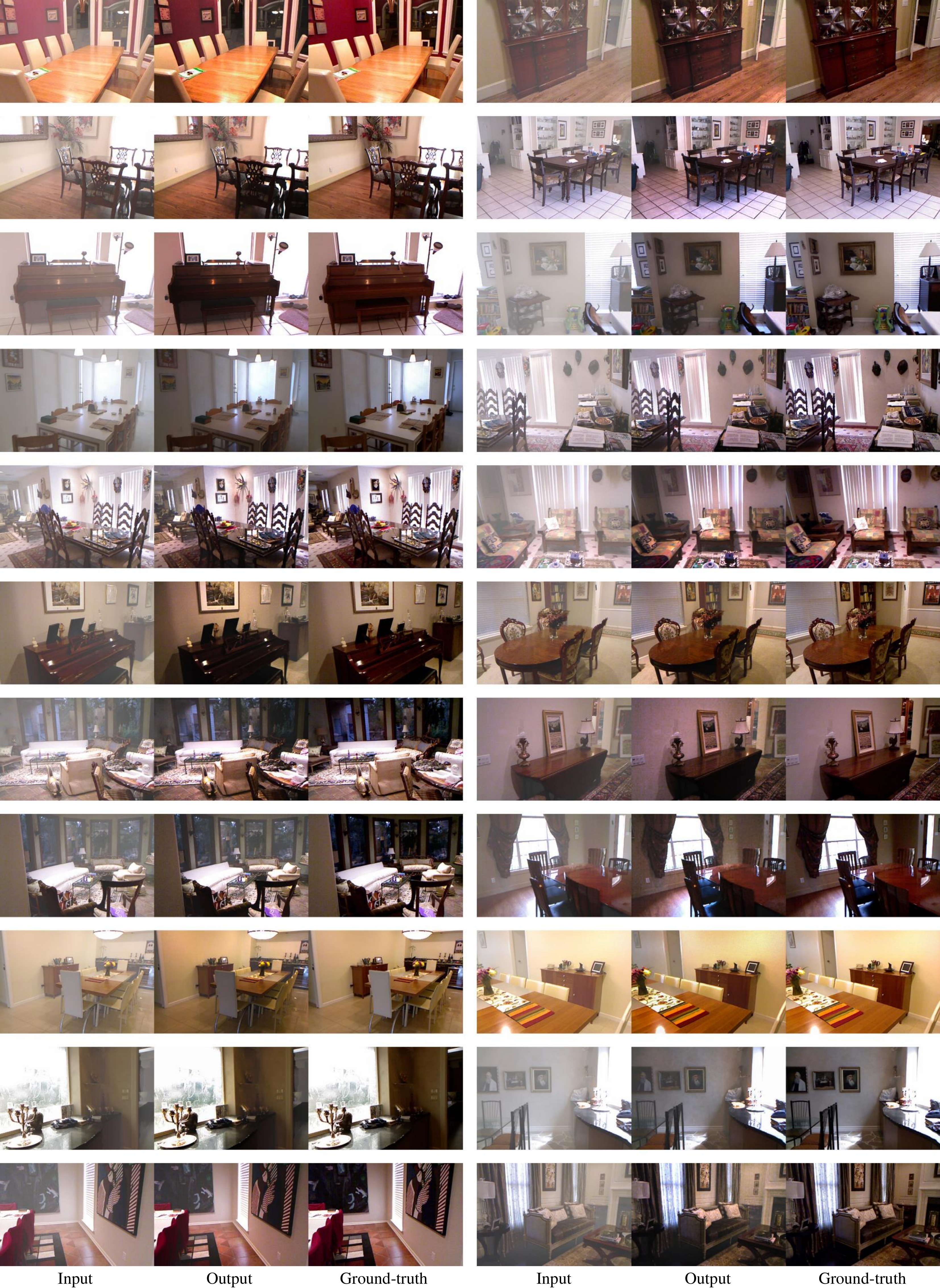}
\caption{Dehzing results on the SOTS-in dataset.} 
\end{figure*}

\begin{figure*}[]
\centering 
\includegraphics[width=1\linewidth]{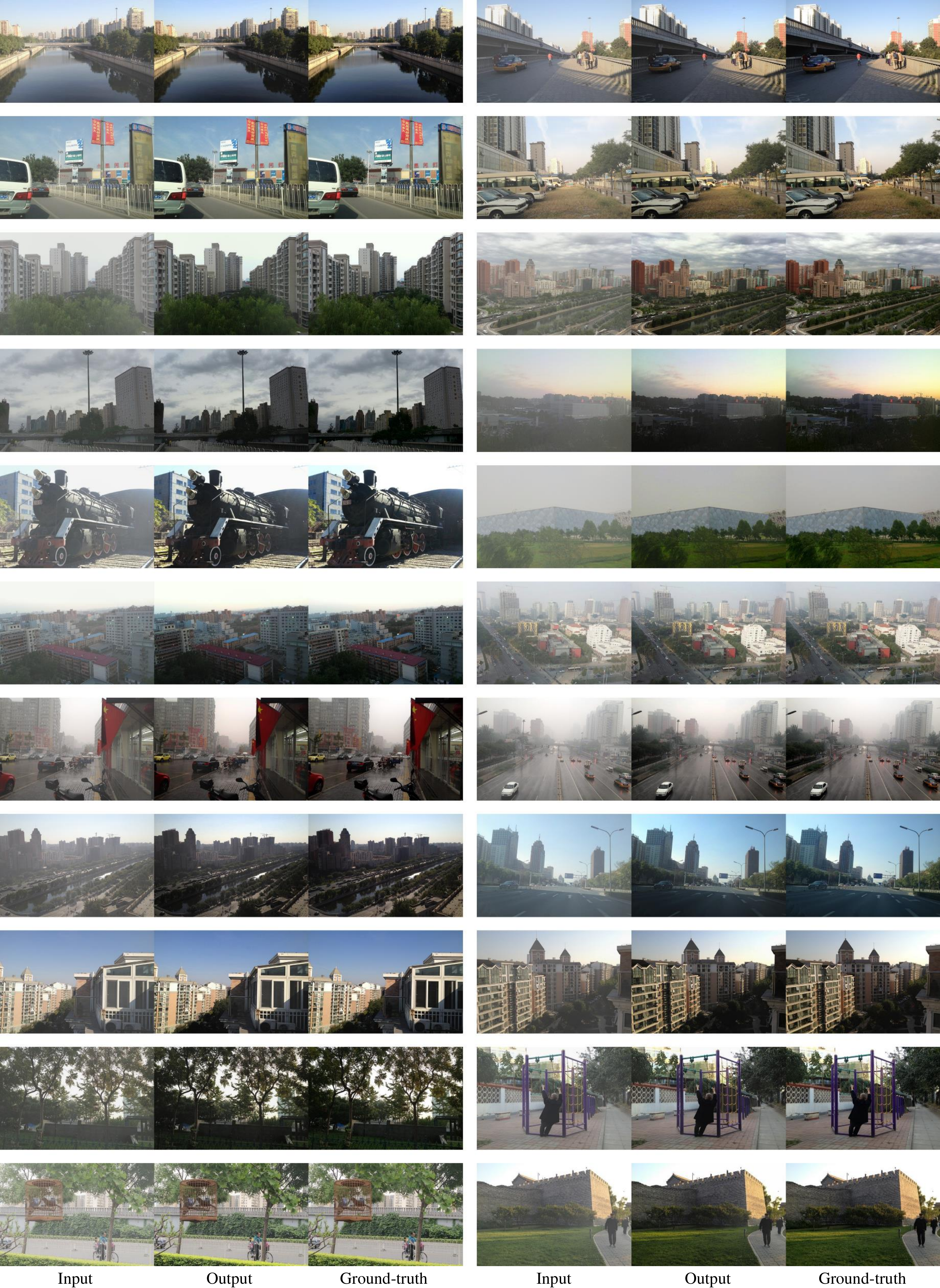}
\caption{Dehazing results on the SOTS-out dataset.} 
\end{figure*}

\clearpage

\subsection{Underwater Image Enhancement}
\begin{figure*}[h]
\centering 
\includegraphics[width=1\linewidth]{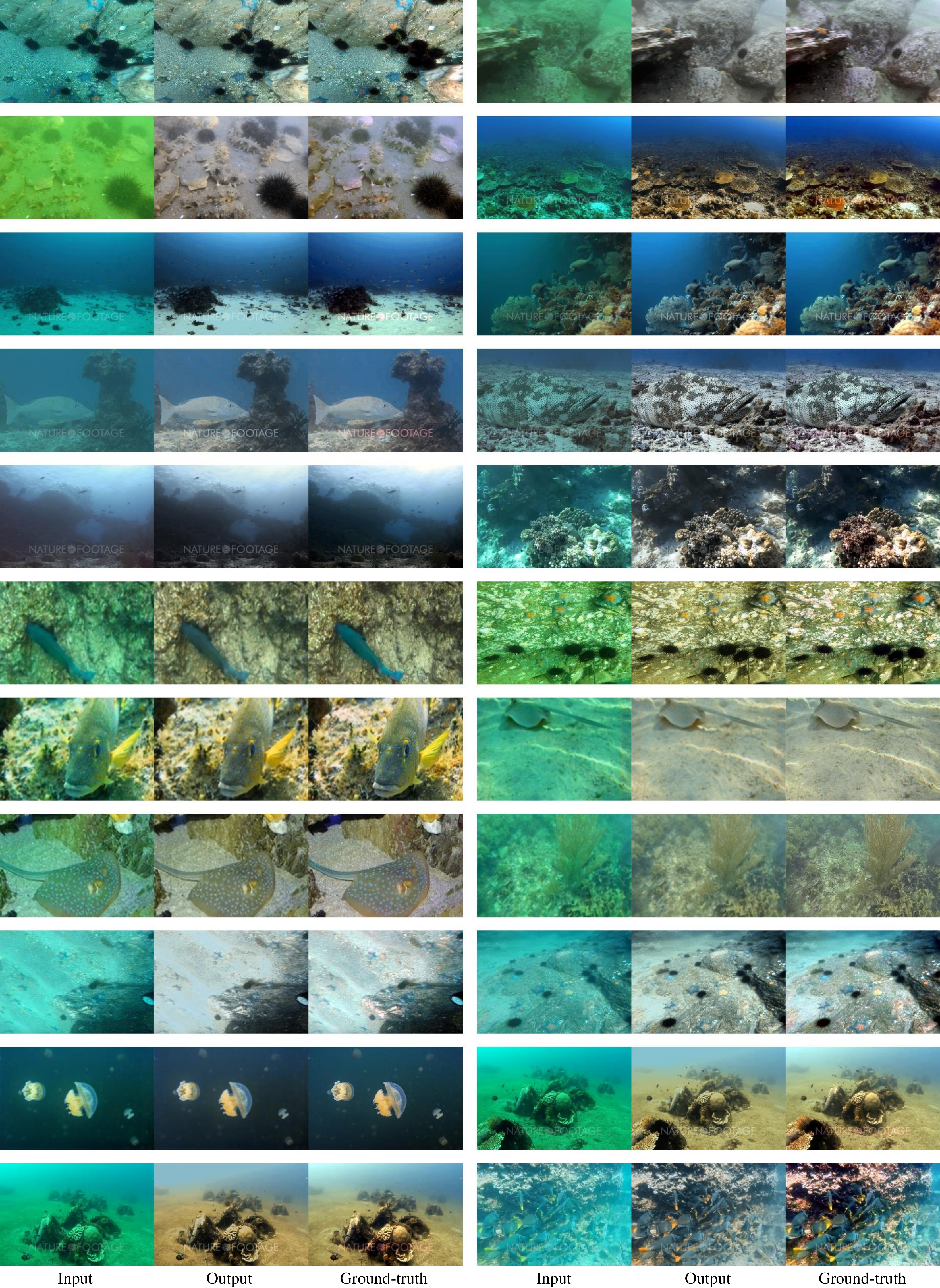}
\caption{UIE results on the LSUI dataset.} 
\end{figure*}

\begin{figure*}[]
\centering 
\includegraphics[width=1\linewidth]{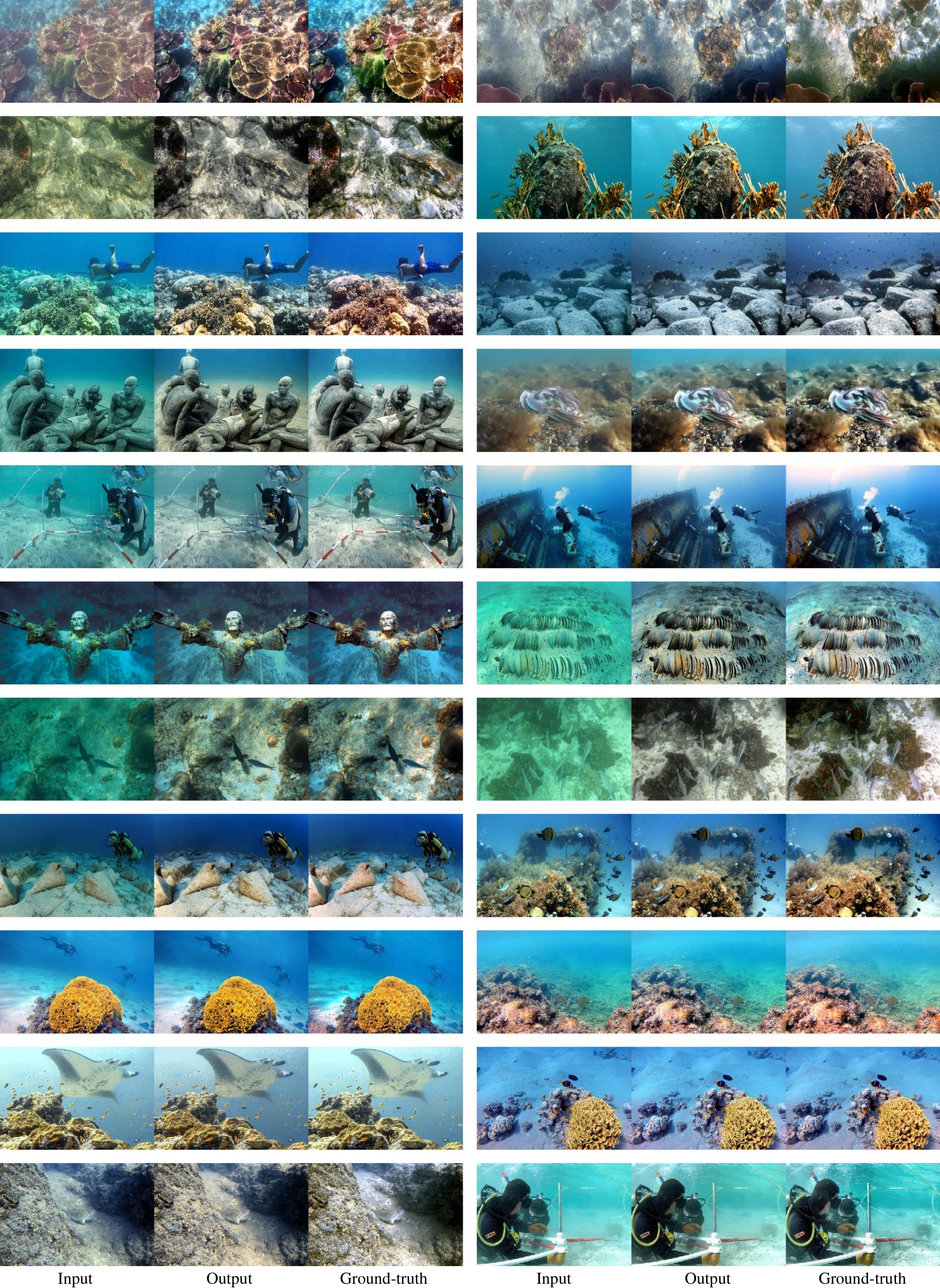}
\caption{Underwater image enhancement results on the UIEB dataset.} 
\end{figure*}

\clearpage 

\subsection{White Balancing}
\begin{figure*}[h]
\centering 
\includegraphics[width=1\linewidth]{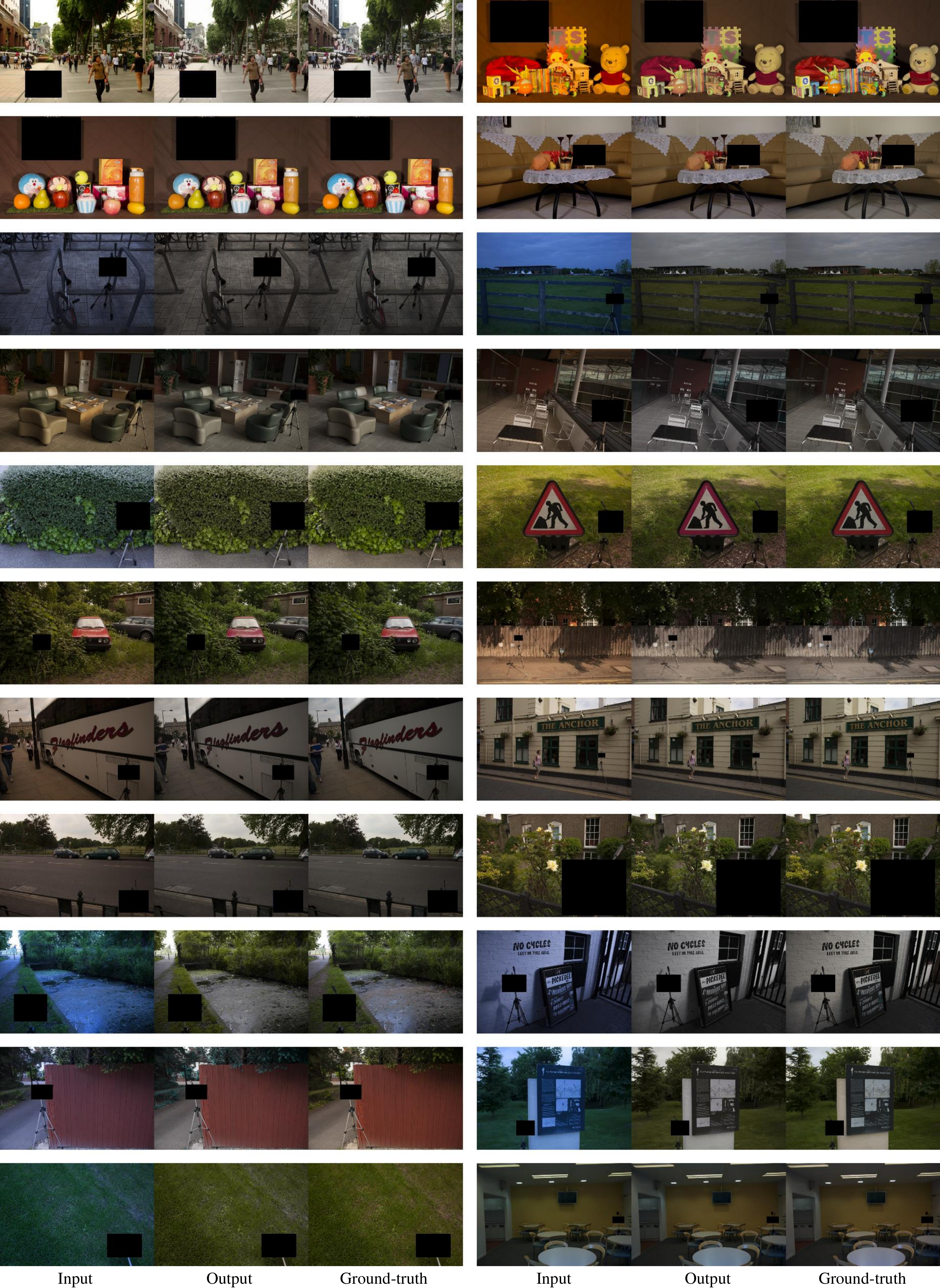}
\caption{WB results on the WB dataset.} 
\end{figure*}

\end{document}